\documentclass[10pt,twocolumn,letterpaper]{article}

\usepackage[pagenumbers]{cvpr}     % To produce the REVIEW version
\usepackage{multirow}
\usepackage{xurl}
\usepackage{xcolor}         % \textcolor

\definecolor{Red}{RGB}{192, 0, 0}
\definecolor{Blue}{RGB}{12, 114, 186} 
\definecolor{Pink}{RGB}{240,56,231} 
\definecolor{myrefcolor}{rgb}{0, 0.367, 0.7}
\definecolor{cvprblue}{rgb}{0.21,0.49,0.74}
\usepackage[pagebackref,breaklinks,colorlinks,allcolors=cvprblue]{hyperref}

\title{Decoupling Action from Egocentric Observation for World Simulation}

\author{
    \textbf{Yue Ma$^{1}$ \quad
    Pengjie Song$^{1}$ \quad
    Xinyu Wang$^{2}$ \quad
    Yi He$^{2}$}
    \\[2pt]
    \textbf{Zeqian Long$^{3}$ \quad
    Fangneng Zhan$^{1}$ \quad
    Kaichen Zhou$^{4}$ \quad
    Hongyu Liu$^{1}$}
    \\[2pt]
    \textbf{Hongfa Wang$^{2}$ \quad
    Peihao Li$^{5}$ \quad
    Haoyang Huang$^{5}$ \quad
    Nan Duan$^{5}$ \quad
    Qifeng Chen$^{1,*}$}
    \\[3pt]
    {\normalfont
    $^{1}$HKUST \quad
    $^{2}$Tsinghua University \quad
    $^{3}$Stanford University  \quad
    $^{4}$MIT \quad
    $^{5}$JD Explore Academy}
    \\[3pt]
    {\normalfont
    \textbf{Project Page: }\href{https://ego-act.github.io/}{https://ego-act.github.io/}}
}
\begin{document}

% ==========================================
% Title + Authors + Teaser (Full Width)
% ==========================================

% \twocolumn[{%
%     \renewcommand\twocolumn[1][]{#1}%
%     \maketitle
%     \vspace{-15pt}
%     \begin{center}
%         \normalfont
%         \normalsize

%         \includegraphics[
%             width=0.9\textwidth
%         ]{images/teaser.pdf}

%         \captionof{figure}{
%             \textbf{Showcase of our method.}
%             Given an egocentric manipulation video as observation,
%             EgoACT decouples the transferable action semantics
%             and generates new interaction videos that faithfully
%             reproduce the reference action while maintaining
%             physically plausible hand-object interactions
%             in the target scene.
%         }
%         \label{fig:teaser}
%     \end{center}

%     % \vspace{0.2cm}
% }]

\twocolumn[{
\renewcommand\twocolumn[1][]{#1}

\maketitle

\vspace{-32pt}

\begin{center}
    \centering
    \includegraphics[width=\linewidth]{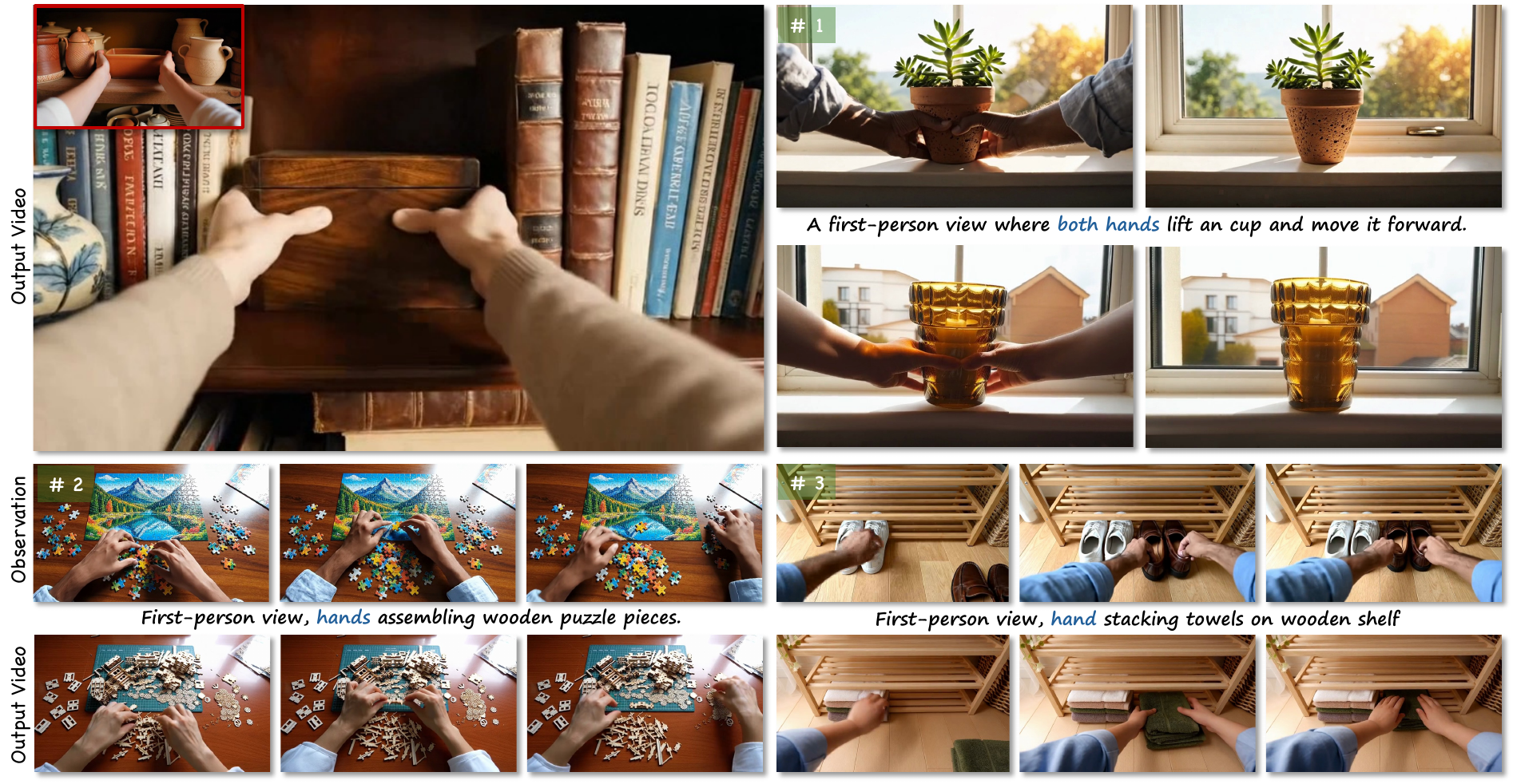}
    \captionsetup{font=normalsize}
    \captionof{figure}{
\textbf{Showcase of our method.}
        Given an egocentric manipulation video as observation,
        EgoACT decouples the transferable action semantics
        and generates new interaction videos that faithfully
        reproduce the reference action while maintaining
        physically plausible hand-object interactions
        in the target scene.
    }
    \label{fig:teaser}
\end{center}
}]

\begingroup
\renewcommand{\thefootnote}{*}
\footnotetext[1]{Corresponding author.}
\endgroup
% ==========================================
% Main Text (Two Columns)
% ==========================================

\begin{abstract}
Egocentric action transfer aims to decouple the semantic action from egocentric observations and reproduce them in new visual contexts, enabling scalable world simulation, embodied policy learning, and robotic data generation. Existing approaches to transferring actions rely on condition-guided video generation, which converts the observation video into explicit geometric controls (\textit{e.g.}, hand poses or meshes) to drive synthesis. However, these methods introduce geometric estimation errors, making it difficult to preserve physically plausible interactions when transferred to new scenes. Alternatively, motion transfer methods directly extract motion patterns from the observation video, yet motion-level features alone cannot encode rich contact dynamics, frequently leading to severe hand structural collapse and implausible contact layout. To address both limitations, we present EgoACT, a test-time framework for egocentric action transfer that operates directly on the denoising process of video diffusion models without requiring additional geometric estimators. EgoACT uses Velocity-guided Structure Anchoring to stabilize reference-consistent hand-object structure in the early denoising stage, and Sparse Correspondence Calibration to refine reliable local correspondences in the mid-to-late denoising stages. Together, these two components preserve transferable action semantics while improving temporal coherence and interaction realism. We further establish EgoActionBench, a benchmark for evaluating action preservation, visual quality, and hand-object plausibility across diverse egocentric manipulation scenarios. Experiments show that EgoACT generates more coherent and physically plausible interaction videos than strong baselines.

\end{abstract}

\section{Introduction}
\label{sec:introduction}

Egocentric action transfer aims to decouple the semantic action from egocentric observations and reproduce it in new visual scenes, enabling scalable world simulation, embodied policy learning, and robotic data generation. The core challenge is to decouple \textit{what action is being performed} from \textit{how the current scene happens to look}, so that the interaction logic, including contact transitions, object responses, and coordinated hand-object dynamics , can be faithfully reproduced in novel scenarios.

% Existing transferable generation methods generally fall into two paradigms: condition-guided video generation~\cite{xing2024make,zhao2023controlvideo,wang2024motionctrl} and general video motion transfer~\cite{pondaven2025video,ma2026fastvmt,ling2024motionclone}. The former converts a observation video into explicit geometric controls (e.g., keypoints, hand poses, or mesh sequences~\cite{ge2018hand,xiao2024spatialtracker,hu2025depthcrafter}) to force the target video to follow, while the latter attempts to extract and replay motion patterns directly from the observation. However, both paradigms struggle when applied to egocentric manipulation. General motion transfer methods often suffer from severe hand structural collapse and lack physically plausible interaction. Condition-guided approaches are similarly ineffective, as egocentric manipulation requires more than just replaying where the hand moves. Hard geometric conditions only weakly encode semantic interactions and are highly sensitive to estimator errors, occlusion, perspective distortion, and hand-object entanglement, often producing pose-aligned but physically rigid interactions.

\begin{figure}[h]%{0.6\linewidth}
  \vspace{-10pt}
  \centering
  \includegraphics[width=\columnwidth]{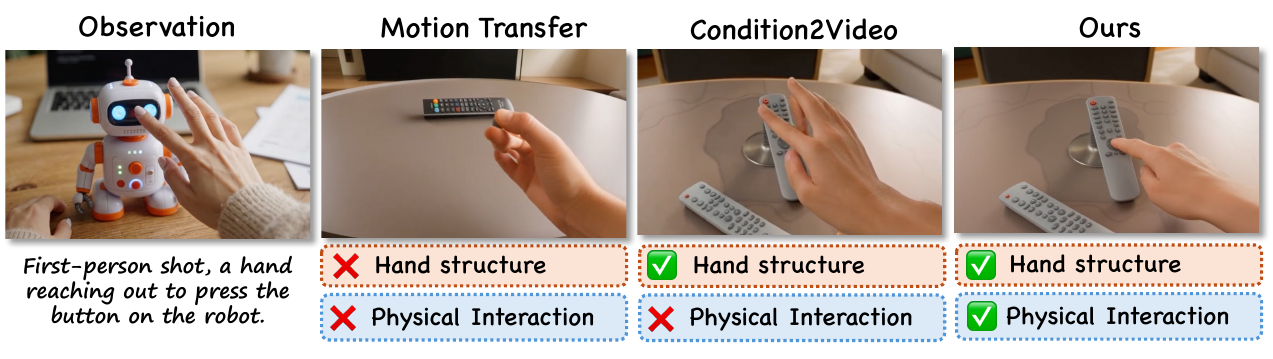}
  \vspace{-20pt}
  \caption{\textbf{Motivation.} Motion transfer suffers from hand structural collapse and lacks physical interaction. Condition-guided generation preserves hand shape but produces physically implausible contact. Our method maintains both hand structure and realistic interaction.}
  \label{fig:intro_motivation}
  \vspace{-10pt}
\end{figure}

Existing approaches to transferring egocentric actions are predominantly formulated as condition-guided video generation~\cite{xing2024make,zhao2023controlvideo,wang2024motionctrl}. These methods first convert an observation video into explicit geometric controls, such as keypoints, hand poses, or mesh sequences~\cite{ge2018hand,xiao2024spatialtracker,hu2025depthcrafter}, and then force the target video to follow these extracted conditions. However, these methods rely on accurate geometric estimation and inevitably introduce external errors into the generation process. When transferred to a new scene, the injected geometric controls cannot guarantee physically valid interactions (Fig.~\ref{fig:intro_motivation}). An alternative line of work, motion transfer~\cite{pondaven2025video,ma2026fastvmt,ling2024motionclone}, sidesteps explicit geometric estimation by extracting and replaying motion patterns directly from the observation video. While this avoids estimator-induced errors, motion-level features alone cannot encode the rich contact dynamics inherent in egocentric manipulation. These methods frequently suffer from severe hand structural collapse and implausible contact layout (Fig.~\ref{fig:intro_motivation}).

To address this, instead of hard-copying sparse geometric structures or replaying raw motion from an observation video, we aim to extract the transferable semantic action from an egocentric observation and use it to synthesize a new interaction video under a different visual realization. This setting asks the model to preserve the \emph{interaction logic} behind the motion, including contact timing, object response, and coherent hand-object coordination, rather than the observation appearance or exact geometric condition.

In this work, we present EgoACT, a test-time action transfer framework for egocentric world simulation. We observe that egocentric action transfer faces two major failure modes during generation. First, \emph{hand-object structural collapse} often occurs in the early denoising stage, where action guidance can distort hands, objects, and contact geometry before a stable interaction layout is formed. Second, \emph{physically unrealistic interaction} can still appear even when visible motion is roughly aligned, because local correspondences, contact timing, and object responses may drift in later denoising stages. To address these challenges, EgoACT combines two complementary test-time refinements. \emph{Velocity-guided Structure Anchoring} (VSA) stabilizes early denoising by preserving reference-consistent hand-object structure, while \emph{Sparse Correspondence Calibration} (SCC) refines mid-to-late denoising by correcting reliable local correspondences around interaction regions.

Beyond the method, we establish EgoActionBench, a benchmark for evaluating egocentric action transfer. EgoActionBench contains 350 egocentric manipulation videos covering diverse interaction categories, object types, hand configurations, and scene layouts. It evaluates not only visual fidelity, but also action preservation, temporal coherence, and hand-object interaction plausibility. Experiments on EgoActionBench show that EgoACT produces more coherent transferred actions and more plausible hand-object interactions than strong baselines, supporting the effectiveness of our formulation. Our contributions are summarized as follows:
\begin{itemize}
    \item We analyze the challenges of egocentric action transfer and propose EgoACT, a test-time action transfer framework that decouples transferable action semantics from observation visual appearance without additional training.
    \item Technically, we introduce \emph{Velocity-guided Structure Anchoring} and \emph{Sparse Correspondence Calibration}, which respectively address hand-object structural collapse and physically unrealistic interaction.
    \item We establish EgoActionBench, a benchmark for evaluating egocentric action transfer comprehensively, and conduct extensive experiments that support the effectiveness of our formulation and method.
\end{itemize}

\section{Related Work}

% 缩减版
\noindent\textbf{Video Motion Transfer.}
Motion transfer aims to synthesize novel videos by preserving reference motion dynamics, requiring effective decoupling of spatial appearance and temporal motion. Early works~\cite{guo2023animatediff, xing2024make, xing2024dynamicrafter, feng2025dit4edit, zhao2023controlvideo, wang2024motionctrl, zhang2025framepainter} rely on explicit control signals, which demand large-scale annotations and substantial computation. Consequently, recent diffusion models shift toward implicit control, employing either parameter-efficient tuning~\cite{zhao2024motiondirector, jeong2024dreammotion, ren2024customize, wu2024motionbooth} or inference-time motion extraction~\cite{hu2024motionmaster, qiu2024tfb, qiu2025duet, qiu2025DBLoss, qiu2026dag, yesiltepe2024motionshop, geyer2023tokenflow, qi2023fatezero, ling2024motionclone, xiao2024video}. However, these frameworks primarily target 3D U-Nets~\cite{blattmann2023align}. As models evolve toward Diffusion Transformers (DiTs)~\cite{peebles2023scalable} with fully coupled 3D attention, traditional methods relying on explicit spatio-temporal separation become difficult to apply. Recent DiT-based adaptations explore attention flow optimization~\cite{pondaven2025video}, parameter-efficient tuning~\cite{abdal2025dynamic}, temporal kernel encoding~\cite{shi2025decouple}, and streamlined extraction~\cite{ma2026fastvmt}. However, these methods still focus primarily on visible motion transfer, while egocentric action transfer requires preserving contact transitions and hand-object dynamics.

\noindent\textbf{Interaction-Centric Video Generation.} 
Recent advances in diffusion have increasingly pushed generative video models toward the paradigm of video world modeling~\cite{brooks2024sora, bruce2024genie, hu2023gaia}, where the goal is to capture not only visual appearance but also the underlying dynamics and interactions of the physical world.  Some works~\cite{li2026mask2iv, zhang2025vhoi, HOIGen} address interactive manipulation video generation by introducing structured interaction signals to improve controllability and physical plausibility. Methods such as Mask2IV~\cite{li2026mask2iv} and VHOI~\cite{zhang2025vhoi} leverage mask trajectories or keypoints to guide fine-grained interactions. Other approaches incorporate more explicit control cues or geometric priors, such as InterDyn~\cite{akkerman2025interdyn}, GenHSI~\cite{li2026genhsi}, and ZeroHSI~\cite{li2024zerohsi}.  EgoSim~\cite{hao2026egosim} and Hand2World~\cite{wang2026hand2world} further introduce egocentric conditioning or structured world representations to enhance interaction realism and consistency. 
Despite these advances, existing methods largely rely on external condition estimators~\cite{ge2018hand, xiao2024spatialtracker, hu2025depthcrafter} to guide the generation of physically consistent interaction videos, which limit practical application. In contrast, we propose EgoACT, which enhances interaction consistency without requiring external estimators, enabling more robust interaction world modeling.

\begin{figure*}[t]
  \centering
  \includegraphics[width=\linewidth]{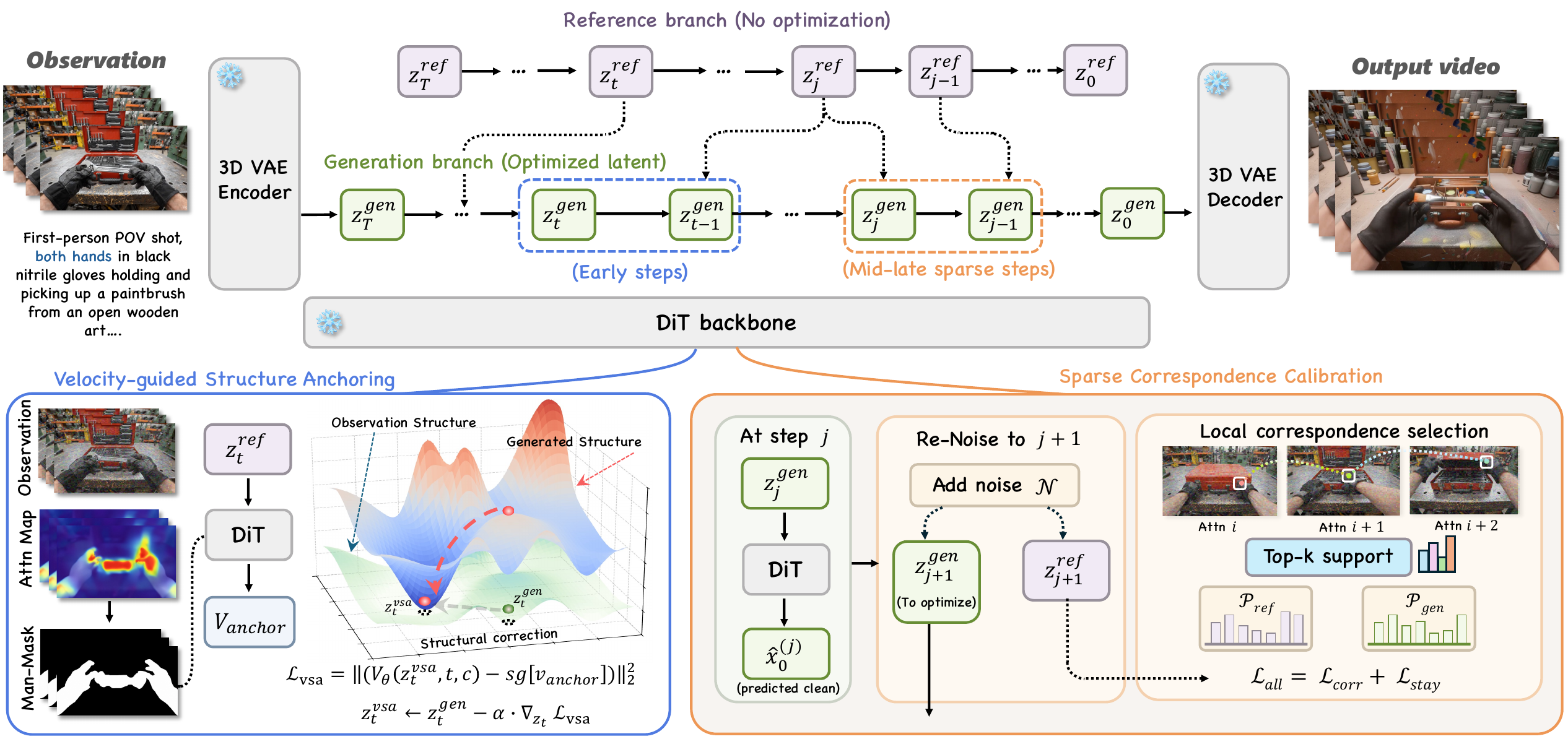}
\caption{\textbf{Overview of the proposed framework.} Given an egocentric manipulation video, our framework uses a reference branch and an optimized generation branch for test-time action transfer. We apply Velocity-guided Structure Anchoring (VSA) in the early denoising steps to preserve hand-object structure, and Sparse Correspondence Calibration (SCC) in the mid-to-late steps to refine local correspondences and improve interaction realism.}
  \label{fig:framework}
\end{figure*}

\section{Method}
\label{sec:method}

Given an egocentric manipulation video $x^{\mathrm{ref}}$ that demonstrates a source action, egocentric action transfer aims to synthesize a new video $x^{\mathrm{gen}}$ that preserves the transferable action semantics while allowing the visual realization to change. The transferred signal should not be limited to visible hand displacement, but should also retain the underlying interaction logic, including contact transitions, object response, and coordinated hand-object dynamics. In this section, we first discuss the motivation and key challenges of egocentric action transfer in Sec.~\ref{subsec:motivation}. We then introduce Velocity-guided Structure Anchoring (VSA) in Sec.~\ref{subsec:VSA}, which stabilizes the global hand-object structure during the early denoising stage. Finally, we present Sparse Correspondence Calibration (SCC) in Sec.~\ref{subsec:scc}, which refines local temporal correspondences in the mid-to-late denoising stages.

\subsection{Motivation}
\label{subsec:motivation}
% 针对两个问题，对应两种方法。
Egocentric action transfer requires the model to transfer the semantic action from a reference egocentric observation while synthesizing a new interaction video under a different visual realization. Consistent with the discussion in Sec.~\ref{sec:introduction}, this setting is difficult because the model must preserve the interaction logic behind the motion rather than simply reproduce visible hand displacement. We focus on two failure modes that commonly appear during generation:

\noindent\textbf{Hand-object Structural Collapse.}
Explicit condition guidance usually converts the reference video into geometric controls, such as keypoints, hand poses, or mesh sequences, and then forces the generation process to follow these controls. Although this can preserve coarse motion, it is fragile in egocentric manipulation, where hands are frequently occluded, distorted by perspective, and tightly coupled with objects. Noisy or distribution-mismatched controls can over-constrain the early denoising trajectory, leading to broken fingers, distorted hands, collapsed objects, or implausible contact geometry.

\noindent\textbf{Physically Unrealistic Interaction.}
Even when visible motion is roughly aligned, the generated video may still fail to preserve the physical logic of manipulation, such as when contact is established, how the object responds, and whether local hand-object correspondences remain temporally consistent. This problem becomes especially apparent in the mid-to-late denoising stages, where fine-grained appearance and contact details are formed.

To address these two failure modes, EgoACT performs two complementary latent optimization steps during inference. Velocity-guided Structure Anchoring (VSA) is applied in the early denoising stage to keep the generation close to reference-consistent hand-object structure, reducing structural collapse before the interaction layout is formed. Sparse Correspondence Calibration (SCC) is then applied in the mid-to-late denoising stages to correct reliable local correspondences around interaction regions, improving temporal consistency and hand-object realism without additional training.

Both modules use a frozen reference branch and an optimized generation branch. At each guided denoising step $t$, the clean reference latent $z_0^{\mathrm{ref}}$ is mapped to the same noise level by the forward process, $z_t^{\mathrm{ref}}=(1-\sigma_t)z_0^{\mathrm{ref}}+\sigma_t\epsilon$, and is used only to extract stop-gradient guidance targets. We update only the generation latent $z_t^{\mathrm{opt}}$. The reference branch is evaluated with a null-text prompt, while the generation branch uses the target prompt.

\subsection{Velocity-guided Structure Anchoring}
\label{subsec:VSA}
% 目前：主要叙述结构的维持。
% 额外的：能不能说Source-Video给了一定的物理知识，然后VSA可以实现对物理知识和结构信息的对齐。
To prevent the hand-object structural layout from being severely derailed during the initial motion injection, we propose \textbf{V}elocity-guided \textbf{S}tructure \textbf{A}nchoring (VSA) at the early guidance steps. In egocentric manipulation, the source observation does not merely provide a motion template; it also encodes implicit physical knowledge about feasible hand-object contacts, relative object motion, and interaction timing. Directly injecting motion cues without preserving this knowledge can produce trajectories that follow the reference displacement but violate the physical structure of the manipulation process. VSA therefore uses the reference branch as a structure-and-physics anchor at the same noise level, aligning the generation trajectory with both the kinematic layout and the interaction dynamics implied by the observation video. 

Specifically, instead of relying purely on spatial loss, we extract the velocity prediction from the reference branch at the same noise level $v_{\mathrm{anchor}} = \operatorname{sg}\bigl[v(z_t^{\mathrm{ref}})\bigr]$, where $V_\theta(\cdot)$ denotes the velocity prediction of the diffusion model parameterized by $\theta$ at noise level $t$ and $\operatorname{sg}[\cdot]$ denotes the stop-gradient operation. Given the velocity prediction $v_{\mathrm{opt}} = v(z_t^{\mathrm{opt}})$ of the currently optimized generation latent, VSA acts as a spatially weighted velocity regression. 

% 这里需要补充，为什么我们结构信息需要用velocity来表示，从流型上描述，大概的思路是，生成结果依旧是当前的flow，而向v_ref对齐，使得结构特征向 z_ref 对齐。
The rationale behind using velocity as the structural carrier is that, in flow-based diffusion sampling, the velocity prediction defines the local tangent direction of the denoising trajectory on the latent video manifold. For egocentric manipulation, this tangent direction implicitly captures how hand poses, object states, and their spatial relations should evolve from the current noisy state. Therefore, aligning $v_{\mathrm{opt}}$ with $v_{\mathrm{anchor}}$ does not directly copy the reference latent; instead, it steers the current generation to follow a reference-consistent local flow. This preserves the freedom of the generated content while encouraging its structural features to move toward the physically plausible manifold indicated by $z_t^{\mathrm{ref}}$.

% 为了更准确的引导，我们引入了AMF的软结构mask。
% 补充对AMF的简要说明。
To avoid applying the velocity anchor uniformly over the entire frame, we introduce an Attention Motion Flow (AMF)~\citep{pondaven2025video} induced soft structural mask $\mathcal M(\mathbf{x})$. The intuition is that regions with stronger reference motion, such as moving hands and manipulated objects, should receive stronger structural anchoring, while static backgrounds should still retain a weak constraint to preserve global coherence. Let $\mathcal A_{\mathrm{ref}}(\mathbf{x})$ denote the reference AMF at spatial location $\mathbf{x}$, and define the structural saliency score as its magnitude:
\vspace{-3pt}
\begin{equation}
\mathcal S(\mathbf{x}) = \|\mathcal A_{\mathrm{ref}}(\mathbf{x})\|_2.
\label{eq:eq1}
\end{equation}
\vspace{-3pt}
We normalize $\mathcal S(\mathbf{x})$ into $[0,1]$ denoted as $\mathcal{\hat S}(\mathbf{x})$ and apply a power transform controlled by $\gamma$ to adjust the concentration of the mask. With a minimum weight bound $m_{\min}$, the final soft mask is formulated as:
\vspace{-3pt}
\begin{equation}
\mathcal{M}(\mathbf{x}) = \mathcal{\hat S}(\mathbf{x})^\gamma (1-m_{\min}) + m_{\min}.
\label{eq:eq2}
\end{equation}
\vspace{-3pt}
The resulting $\mathcal{M}(\mathbf{x})$ is defined only over spatial locations and is shared across all frames during VSA optimization.
% 最后总结内容
Using soft masks and velocity anchors provides structural guidance. The final VSA objective is a mask-weighted velocity regression over $N$ spatial-temporal latent positions:
% \vspace{-3pt}
\begin{equation}
\mathcal{L}_{\mathrm{VSA}}=\frac{1}{N}\sum_{\mathbf{x}}\mathcal{M}(\mathbf{x})\left\|v_{\mathrm{opt}}(\mathbf{x}) - v_{\mathrm{anchor}}(\mathbf{x})\right\|_2^2 .
\label{eq:eq3}
\end{equation}
% \vspace{-3pt}
During early sampling, by gradient descent $z_t^{\mathrm{opt}} \leftarrow z_t^{\mathrm{opt}} - \alpha \nabla_{z_t^{\mathrm{opt}}} \mathcal{L}_{\mathrm{VSA}}$, the latent is directly updated, effectively stabilizing the global structural layout.
% During early sampling, by gradient descent $z_t^{\mathrm{opt}} \leftarrow z_t^{\mathrm{opt}} - \eta \nabla_{z_t^{\mathrm{opt}}} (\mathcal{L}_{\mathrm{core}} + \omega_{\mathrm{VSA}} \mathcal{L}_{\mathrm{VSA}})$, the latent is directly updated, effectively stabilizing the global structural layout.

\subsection{Sparse Correspondence Calibration}
\label{subsec:scc}
% 目前：主要叙述对Ego轨迹的一致性保持，实现可迁移。
% 叙说VSA在轨迹对应上的不足，但当前全局优化的局限性。
While VSA stabilizes the macroscopic structure in the early stage of denoising, cross-frame correspondences may still drift during the following denoising process. Existing global optimization strategies, such as test-time optimization~\cite{kim2025test,li2025test,wu2024self}, usually regularize the noisy latent with dense reconstruction, feature, or motion-level objectives. In egocentric manipulation, however, the visual content and action semantics are tightly coupled: hands, objects, and contact states jointly determine whether a trajectory remains valid. Global objectives tend to dilute these fine-grained vision-action constraints into a single optimization target, which may over-correct the latent and unintentionally alter object appearance, hand-object relations, or the intended action.

To address this, we introduce \textbf{S}parse \textbf{C}orrespondence \textbf{C}alibration (SCC), which operates within a valid diffusion path and aligns the local correspondence distributions between the reference and generated branches at equivalent noise levels. By optimizing only the reliable sparse matching support induced by the reference branch, SCC provides a more targeted correction signal for local interaction-level correspondence drift while avoiding unnecessary updates on ambiguous or irrelevant regions.

% 叙说sparse correspondence的构建方式
Egocentric manipulation videos exhibit locally continuous motion between adjacent frames: a hand or object token is unlikely to jump to an arbitrary global position within one denoising step. Based on this observation, SCC searches for correspondences only within a local displacement window $\mathcal{N}(u)=\{u+\Delta \mid \Delta\in[-r,r]^2\}$ around each query token $u$. Specifically, for each adjacent frame pair $(i,i+1)$, SCC uses the query token from the current frame and the key token from the next frame, i.e., $q_i(u)$ and $k_{i+1}(u+\Delta)$, to measure cross-frame correspondence. In the dual branch, we score each candidate displacement by the scaled query-key similarity:
\vspace{-3pt}
\begin{equation}
s_i(u,\Delta)=\frac{\langle q_i(u), k_{i+1}(u+\Delta) \rangle}{\sqrt{d}},
\quad u+\Delta\in\mathcal{N}(u).
\end{equation}
\vspace{-3pt}
We then keep the top-$k$ displacements with the largest reference scores as $\Omega_k^i(u)$. This reference-induced support specifies where reliable local matches are expected to occur. Both the reference and generated correspondence distributions are normalized over the same support:
\vspace{-3pt}
\begin{equation}
\mathcal P_i(\Delta\mid u) = \operatorname{softmax}\left(\frac{s_i(u,\Delta)}{\tau}\right), \quad \Delta\in\Omega_k^i(u).
\end{equation}
\vspace{-3pt}
Recognizing that not all reference tokens provide reliable matching cues, we assign a confidence weight $w_{\mathrm{conf}}^i(u) = 1 - \frac{H_i(u)}{\log k}$ based on the entropy $H_i(u)$ of the reference distribution, and normalize it as $\bar w_i(u)=w_{\mathrm{conf}}^i(u)/\sum_{u'}w_{\mathrm{conf}}^i(u')$. The correspondence alignment is then supervised by a confidence-weighted cross-entropy loss:
\vspace{-3pt}
\begin{equation}
\begin{aligned}
\mathcal{L}_{\mathrm{corr}}
&= -\sum_i \sum_u \bar w_i(u) \\
&\quad \cdot \sum_{\Delta\in\Omega_k^i(u)}
\mathcal{P}_{\mathrm{ref},i}(\Delta\mid u)
\log \mathcal{P}_{\mathrm{gen},i}(\Delta\mid u).
\end{aligned}
\label{eq:corr_loss}
\end{equation}
\vspace{-3pt}
This objective encourages the generated branch to place high probability on the same local displacements preferred by the reference branch. The confidence weight further makes sharp and reliable reference correspondences dominate the calibration, while suppressing ambiguous tokens with high-entropy matching distributions. In this way, SCC optimizes the cross-frame matching structure rather than directly pulling the latent toward a dense visual target, which helps correct local correspondence drift while preserving the generated content.

% 引入stay损失
To restrict the correction magnitude and prevent the latent from deviating too far from the original sampling path, a stay regularizer $L_{\mathrm{stay}} = \| z_{j+1}^{\mathrm{opt}}-\operatorname{sg}[z_{j+1}] \|_2^2$ is added. The total objective is:
\vspace{-3pt}
\begin{equation}
\mathcal L_{\mathrm{SCC}} = \lambda_{\mathrm{c}}\mathcal L_{\mathrm{corr}} + \lambda_{\mathrm{s}}\mathcal L_{\mathrm{stay}}.
\end{equation}
\vspace{-3pt}
By minimizing $L_{\mathrm{SCC}}$, the model efficiently rectifies shape drift in interactive regions while regularizing the latent to stay close to the original sampling path.

\begin{figure*}[t]
  \centering
  \includegraphics[width=0.9\linewidth]{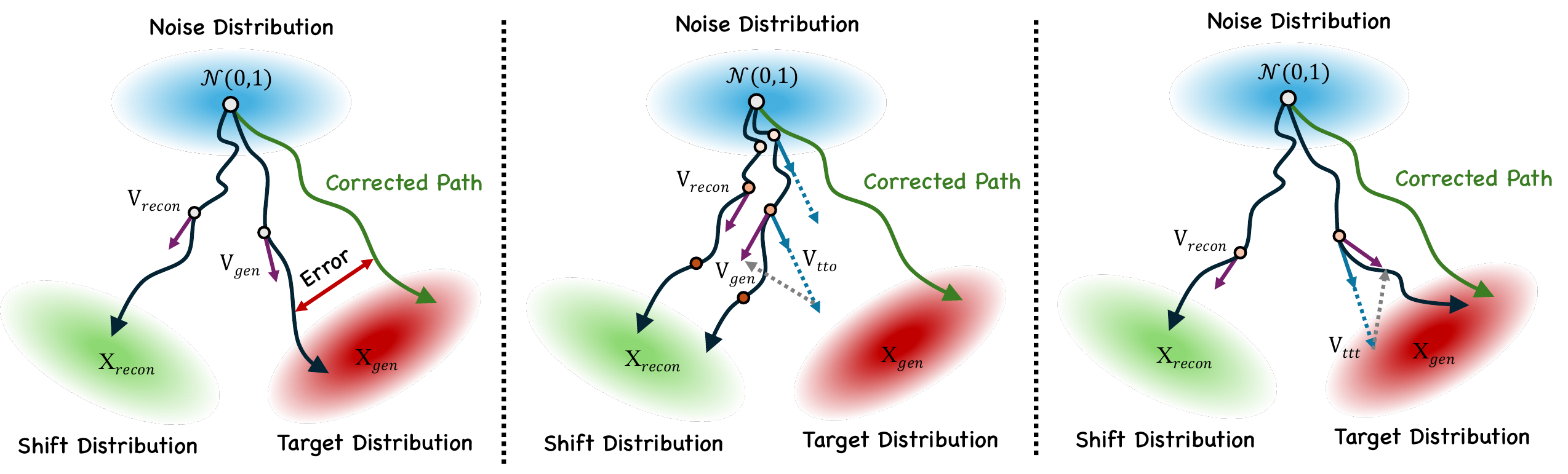}
  \caption{\textbf{Illustration of the proposed SCC.} (a) Direct transfer injects the reference action without correction, which may lead to errors in weak hand-object interactions. (b) Applying correction throughout the whole denoising process over-constrains early generation, making it difficult to decouple transferable action from source observation. (c) Our SCC performs sparse correspondence calibration only in the mid-to-late denoising stages, where the action layout has been established and local interaction details can be refined, improving temporal consistency and hand-object realism.}
  \label{fig:scc_motivation}
\end{figure*}

% 可能画的Motivation：
% 为什么VSA？流型可学习。但这一部分在前面的图中有过了。
% 为什么SCC？VSA无法关注局部错配，现有的TTO全局优化。

% 可能画的Figure：
% 1. AMF -> 软结构mask
% 2. 相邻帧变化不大 -> \deta的可行性。
% 3. L_corr的可视化，结合信息熵。

\section{Experiments}

\subsection{Implementation Details}
In our experiments, we implement EgoACT on top of a pre-trained video diffusion transformer~\cite{wan2025wan} and keep all model parameters frozen during inference. Unless otherwise specified, videos are sampled with 50 denoising steps at a resolution of \textcolor{black}{832}$\times$\textcolor{black}{480} and \textcolor{black}{81} frames. During inference, we apply the sliding-window AMF guidance in the first 20\% of the outer denoising process to stabilize the hand-object layout, and use SCC in the mid-to-late denoising stage from step \textcolor{black}{35} to \textcolor{black}{45} to calibrate local correspondences. Each guided outer step performs a 10-step latent-only optimization with AdamW, where the learning rate is linearly decayed from 0.003 to 0.002. More implementation details are provided in the supplementary material.

\begin{figure*}[t]
  \centering
  \includegraphics[width=0.9\linewidth]{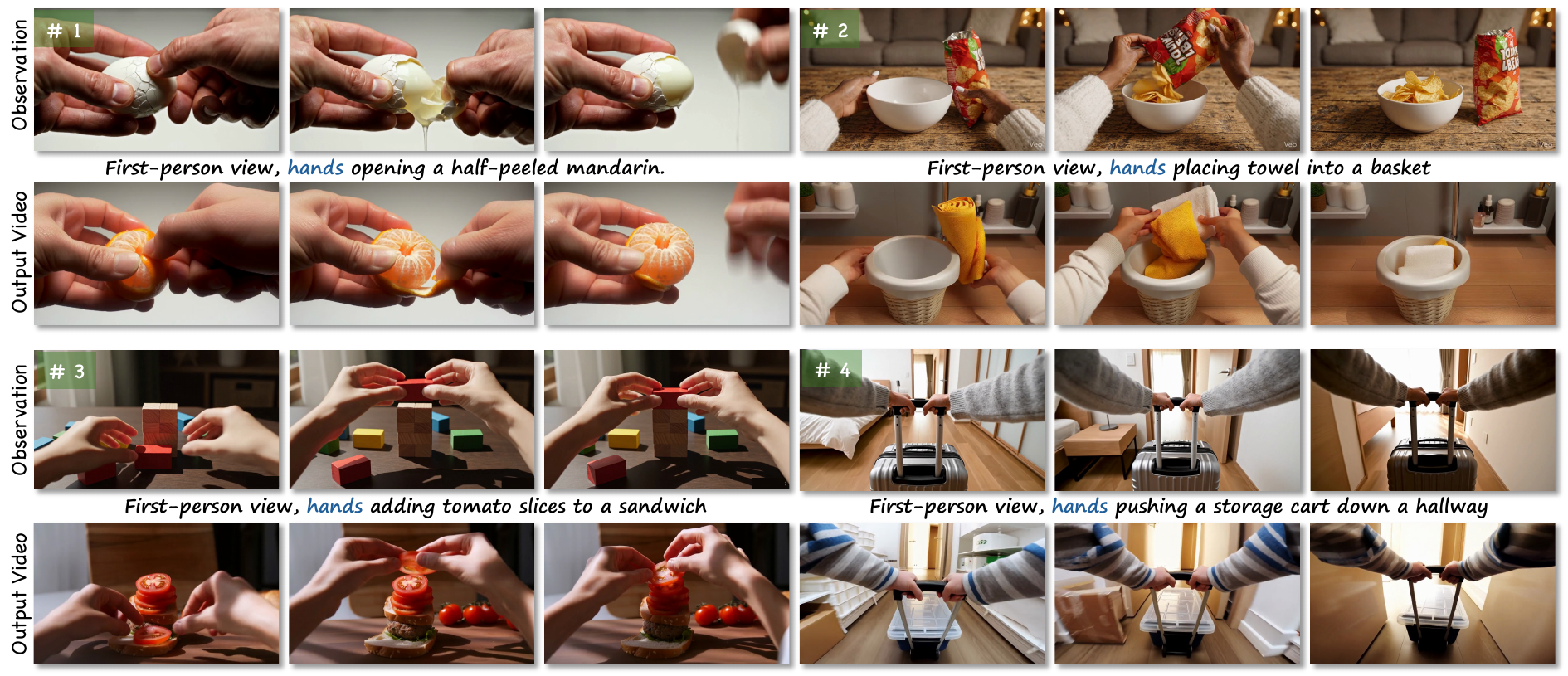}
  \caption{\textbf{Gallery of proposed method.} EgoACT can transfer a wide range of egocentric manipulation videos, covering diverse interaction categories, object types, and motion patterns. The generated videos preserve plausible hand-object structure, contact dynamics, and temporal coherence while adapting the reference action to new visual contexts.}
  \label{fig:gallery}
\end{figure*}

\begin{table*}[t]
  \centering
  \caption{\textcolor{black}{\textbf{Comparison with state-of-the-art methods}.   
  \textcolor{Red}{\textbf{Red}} and \textcolor{Blue}{\textbf{Blue}} denote the best and second best results, respectively.}}
  \label{tab:comparison}
  \resizebox{\textwidth}{!}{%
    \begin{tabular}{l|cccc|cccc}
      \toprule
      \multirow{2}{*}{Method}
        & \multicolumn{4}{c|}{Quantitative Metrics}
        & \multicolumn{4}{c}{Vbench Metrics} \\
      \cmidrule(lr){2-5}\cmidrule(lr){6-9}
        & Text Sim.$\uparrow$ & Motion Fid.$\uparrow$ & Temp. Cons.$\uparrow$ & App. Cons.$\uparrow$
        & Sub. Cons.$\uparrow$ & Back. Cons.$\uparrow$ & Aes. Qual.$\uparrow$ & Motion Smooth.$\uparrow$ \\
        \midrule
        InterDyn~\citep{akkerman2025interdyn}
        & 0.2645 & \textcolor{Blue}{\textbf{0.2434}} & 0.9673 & \textcolor{Blue}{\textbf{0.9265}}
        & 0.9105 & 0.9069 & 0.4273 & 0.9865 \\
      Mask2IV~\citep{li2026mask2iv}
        & 0.2604 & 0.1910 & 0.9749 & 0.8565
        & 0.8929 & 0.9495 & 0.4673 & 0.9820 \\
      CosHand~\citep{sudhakar2024controlling}
        & 0.2602 & 0.2227 & 0.9277 & 0.9133
        & 0.8381 & 0.9201 & 0.4226 & 0.8784 \\
      DiTFlow~\citep{pondaven2025video}
        & 0.2579 & 0.2167 & \textcolor{Blue}{\textbf{0.9750}} & 0.8839
        & 0.8856 & 0.9325 & \textcolor{Blue}{\textbf{0.5065}} & 0.9868 \\
      VACE~\citep{jiang2025vace}
        & \textcolor{Blue}{\textbf{0.2624}} & 0.2389 & 0.9571 & 0.9149
        & \textcolor{Blue}{\textbf{0.9439}} & \textcolor{Blue}{\textbf{0.9507}} & 0.4647 & \textcolor{Blue}{\textbf{0.9881}} \\
        \midrule
      Ours
        & \textcolor{Red}{\textbf{0.2746}} & \textcolor{Red}{\textbf{0.2750}} & \textcolor{Red}{\textbf{0.9866}}   & \textcolor{Red}{\textbf{0.9617}}
        & \textcolor{Red}{\textbf{0.9732}} & \textcolor{Red}{\textbf{0.9719}} & \textcolor{Red}{\textbf{0.5633}} & \textcolor{Red}{\textbf{0.9898}} \\
      \bottomrule
    \end{tabular}%
}
\end{table*}

\begin{figure*}[t]
   \centering
   \includegraphics[width=0.9\linewidth]{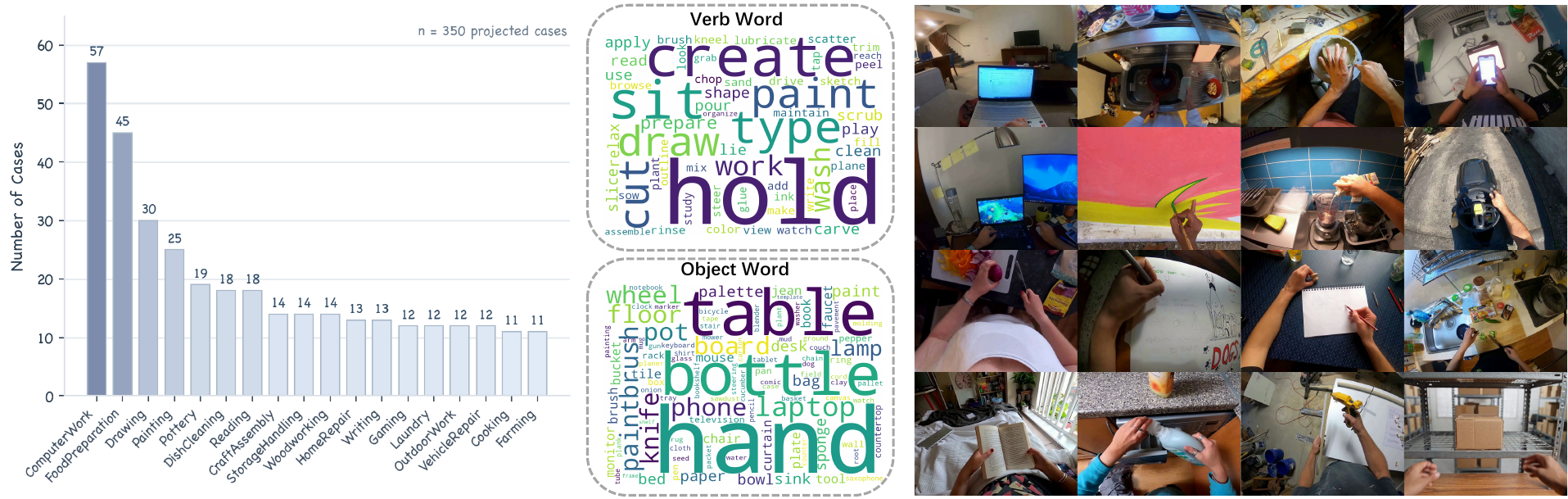}
   \caption{\textbf{The overview of EgoActionBench.} EgoActionBench contains 350 reference manipulation videos and 1,400 target prompts, covering indoor/outdoor scenes, single-/two-hand interactions, and diverse manipulation categories. The benchmark evaluates whether generated videos preserve the reference action semantics while adapting to new visual contexts.}
   \label{fig:egoactionbench}
\end{figure*}

\noindent \textbf{EgoActionBench.} We establish EgoActionBench to evaluate egocentric action transfer in action-centric world simulation.
EgoActionBench is built from online egocentric manipulation videos, public manipulation datasets~\citep{genrobot2025ego, hoque2025egodex, genrobot2025realomni}, and generated videos from powerful video diffusion models~\citep{wan2025wan, hong2022cogvideo,kong2024hunyuanvideo}. It contains 350 high-quality observation videos and 1,400 evaluation prompts, covering 287 indoor and 63 outdoor scenes, with 164 single-hand and 186 two-hand interactions. The actions include grasping, opening, pouring, placing, tool use, and handover across diverse objects and interaction patterns. To ensure the visual quality of benchmark, we filter out clips with excessive ego-camera motion, severe blur, heavy occlusion, objects leaving the frame, unclear hand-object contact, incomplete actions, or corrupted frames. For each retained video, GPT-5 generates an action-centric source caption, from which we derive 4 target prompts by changing the scene, object appearance, or background while preserving the original action semantics. All prompts are manually checked to avoid reference-specific appearance leakage. Each sample is annotated with scene type, hand type, action category, and object category.

\subsection{Comparison with Baselines}

\begin{figure}
   \vspace{-10pt}
   \centering
   \includegraphics[width=\linewidth]{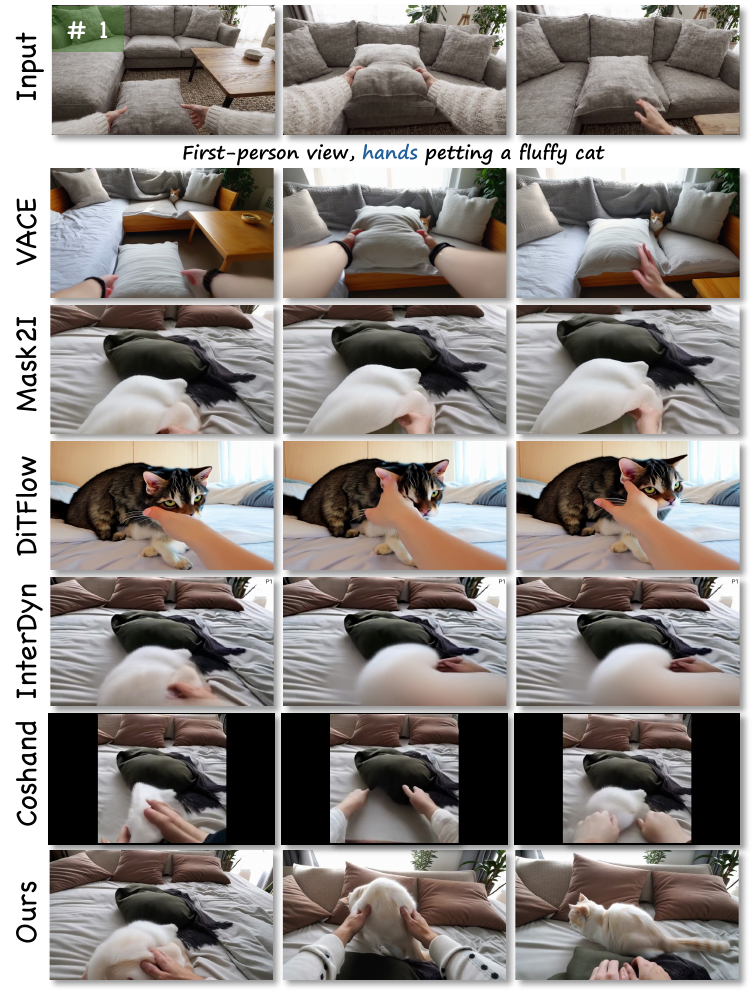}
   \caption{\textbf{Qualitative comparison.} Compare with previous methods, our method preserves stable hand-object layout and object response.}
   \label{fig:qualitative_comparison}
\end{figure}

\noindent \textbf{Qualitative comparison.} We compare EgoACT with representative egocentric manipulation and interaction generation methods, including VACE~\citep{jiang2025vace}, Mask2IV~\citep{li2026mask2iv}, CosHand~\citep{sudhakar2024controlling}, InterDyn~\citep{akkerman2025interdyn}, and DiTFlow~\citep{pondaven2025video}. As shown in Fig.~\ref{fig:qualitative_comparison}, previous methods exhibit different failure patterns in egocentric manipulation transfer. Mask2IV and CosHand better preserve hand shape but struggle with contact consistency, often generating implausible object responses such as objects remaining static despite clear grasping motion. InterDyn and DiTFlow introduce visible structural artifacts when hands and objects are tightly coupled, leading to floating contacts or broken spatial relations. In contrast, EgoACT preserves a more stable hand-object layout while maintaining temporally coherent contact and physically plausible object response throughout the interaction sequence. These qualitative results confirm that VSA suppresses early structural collapse by anchoring the hand-object skeleton, while SCC improves local correspondence consistency in the later denoising stages to produce more realistic contact dynamics.
\begin{figure*}
  \centering
  \includegraphics[width=0.95\linewidth]{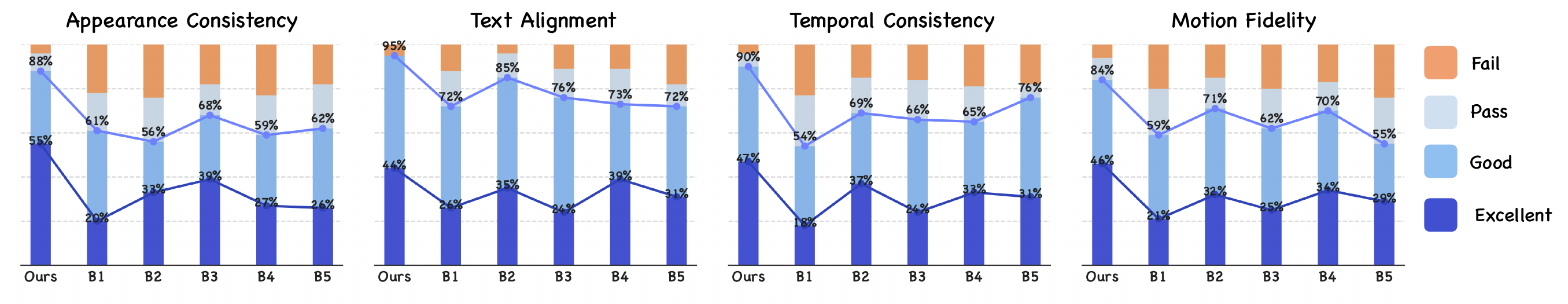}
  \caption{\textbf{User Study.} B1--B5 denote VACE, Mask2IV, CosHand, InterDyn, and DiTFlow, respectively. Our method is consistently preferred by human evaluators, achieving the highest excellent/good ratio across all evaluated aspects.}
  \label{fig:user_study}
\end{figure*}

\noindent \textbf{Quantitative comparison.}
We quantitatively compare EgoACT with VACE~\citep{jiang2025vace}, Mask2IV~\citep{li2026mask2iv}, CosHand~\citep{sudhakar2024controlling}, InterDyn~\citep{akkerman2025interdyn}, and DiTFlow~\citep{pondaven2025video} on EgoActionBench. For fair comparison, all methods are evaluated on the same set of reference manipulation videos and target prompts. Following prior works, we adopt four task-level metrics: (a) \emph{Text Similarity}, the average CLIP similarity between the target prompt and generated frames; (b) \emph{Motion Fidelity}, the tracklet similarity between reference and generated videos; (c) \emph{Temporal Consistency}, the average feature similarity between consecutive frames; and (d) \emph{Appearance Consistency}, measuring how well the generated video preserves the target visual appearance.

\noindent \textbf{User Study.}
To account for the limitations of automatic metrics in capturing real-world preferences, we conduct a user study with 20 volunteers on 50 randomly selected test cases. The volunteers categorize each result into one of four levels: excellent, good, pass, or fail, based on appearance consistency, text alignment, temporal consistency, and motion fidelity. In Fig.~\ref{fig:user_study}, B1--B5 correspond to VACE, Mask2IV, CosHand, InterDyn, and DiTFlow, respectively. Our method outperforms others in both automated metrics and human subjective preferences.

\noindent \textbf{VBench Metrics.} We further adopt four VBench metrics for a broader assessment of video quality. \emph{Subject Consistency} evaluates whether the main subject identity is preserved across frames. \emph{Background Consistency} measures the coherence of static scene regions. \emph{Aesthetic Quality} scores the visual appeal of generated videos, and \emph{Motion Smoothness} evaluates inter-frame motion continuity. As shown in Table~\ref{tab:comparison}, EgoACT achieves strong performance across all metrics, which further validates the effectiveness of our method in producing high-quality, coherent, and semantically aligned egocentric manipulation videos.

% \vspace{-1pt}

\subsection{Ablation Study}
\label{sec:Ablation Study}
We ablate the two core components of EgoACT under the same experimental setup as the main comparison to validate their individual contributions.

\begin{table}[t]
  \centering
  \renewcommand{\arraystretch}{1.1}
  \caption{\textbf{Quantitative ablation.}
  Red and blue denote the best and second-best results, respectively.}
  \label{tab:ablation}
  % \vspace{-1mm}
  {\scriptsize
  \begin{tabular}{@{}l|cccc@{}}
    \toprule
    Method
      & Text Sim.$\uparrow$
      & Motion Fid.$\uparrow$
      & Temp. Cons.$\uparrow$
      & App. Cons.$\uparrow$ \\
    \midrule
    W/o VSA
      & \textcolor{Blue}{\textbf{0.2689}}
      & \textcolor{Blue}{\textbf{0.2442}}
      & 0.9683
      & 0.9312 \\
    W/o SCC
      & 0.2641
      & 0.2398
      & \textcolor{Blue}{\textbf{0.9714}}
      & \textcolor{Blue}{\textbf{0.9487}} \\
    \midrule
    Ours
      & \textcolor{Red}{\textbf{0.2746}}
      & \textcolor{Red}{\textbf{0.2750}}
      & \textcolor{Red}{\textbf{0.9866}}
      & \textcolor{Red}{\textbf{0.9617}} \\
    \bottomrule
  \end{tabular}
  }
  % \vspace{-3mm}
\end{table}

\begin{figure}[t]
  \centering
  \includegraphics[width=\linewidth]{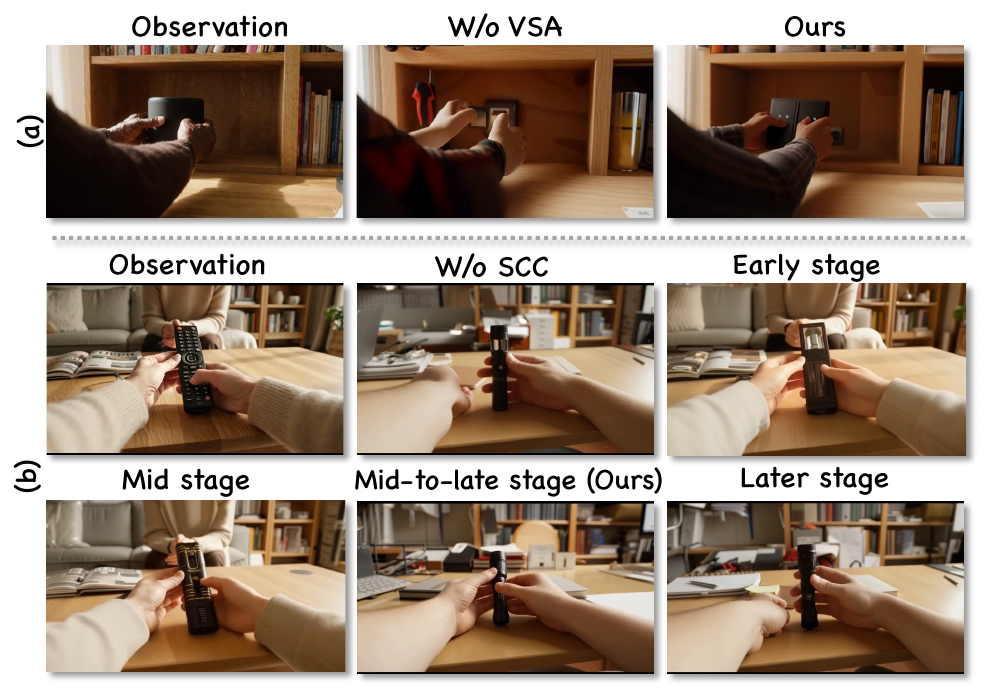}
  % \vspace{-2mm}
  \caption{\textbf{Visual ablation.}
  VSA suppresses structural collapse, while SCC improves local correspondence.}
  \label{fig:ablation}
  % \vspace{-4mm}
\end{figure}

\noindent\textbf{Effectiveness of VSA.}
As shown in Table~\ref{tab:ablation} and Fig.~\ref{fig:ablation}, removing VSA leads to notable degradation in hand generation quality. Without velocity-guided anchoring in early denoising, the generation trajectory lacks structural constraints from the reference branch, causing frequent hand distortion, broken fingers, and implausible contact geometry. With VSA enabled, the reference velocity field steers the latent toward a structurally consistent manifold, preventing structural collapse before the interaction skeleton is established.

\noindent\textbf{Effectiveness of SCC.}
Disabling SCC preserves coarse hand motion but degrades fine-grained interaction quality. Without correspondence calibration, local hand-object correspondences drift across frames, resulting in contact jitter and delayed object response (Table~\ref{tab:ablation}, Fig.~\ref{fig:ablation}). We apply SCC in the mid-to-late denoising stages (steps 35--45): early-stage correction over-constrains the latent and disrupts action-appearance decoupling, while late-stage intervention has limited effect as fine details have solidified. This window allows SCC to refine correspondences after the global layout is stabilized by VSA, yielding more plausible interactions.

\section{Conclusion}

We present EgoACT, a test-time action transfer framework for egocentric world simulation that decouples transferable action semantics from source visual appearance without additional training. EgoACT addresses two critical failure modes in egocentric manipulation transfer: hand-object structural collapse in early denoising and physically unrealistic interaction in later stages. Velocity-guided Structure Anchoring (VSA) stabilizes the global hand-object layout by aligning the generation trajectory with reference-consistent velocity fields during early denoising, while Sparse Correspondence Calibration (SCC) refines local correspondences around interaction regions in the mid-to-late stages to improve contact consistency and object response plausibility. We also introduce EgoActionBench, a benchmark with 350 egocentric manipulation videos and 1,400 evaluation prompts spanning diverse scenes, hand configurations, and interaction categories. Extensive experiments demonstrate that EgoACT outperforms existing methods in both quantitative metrics and human evaluation, producing more coherent action transfer with plausible hand-object interactions.

{
    \small
    \bibliographystyle{ieeenat_fullname}
    \bibliography{main}

@article{brooks2024sora,
  title={Video generation models as world simulators},
  author={Brooks, Tim and Peebles, Bill and Holmes, Connor and DePue, Will and Guo, Yufei and Jing, Leo and Schnurr, David and Taylor, Joe and Luhman, Troy and Luhman, Eric and others},
  journal={OpenAI Blog},
  volume={1},
  number={8},
  pages={1},
  year={2024}
}

@inproceedings{bruce2024genie,
  title={Genie: Generative interactive environments},
  author={Bruce, Jake and Dennis, Michael D and Edwards, Ashley and Parker-Holder, Jack and Shi, Yuge and Hughes, Edward and Lai, Matthew and Mavalankar, Aditi and Steigerwald, Richie and Apps, Chris and others},
  booktitle={Forty-first International Conference on Machine Learning},
  year={2024}
}

@article{hoque2025egodex,
  title={Egodex: Learning dexterous manipulation from large-scale egocentric video},
  author={Hoque, Ryan and Huang, Peide and Yoon, David J and Sivapurapu, Mouli and Zhang, Jian},
  journal={arXiv preprint arXiv:2505.11709},
  year={2025}
}

@inproceedings{jiang2025vace,
  title={Vace: All-in-one video creation and editing},
  author={Jiang, Zeyinzi and Han, Zhen and Mao, Chaojie and Zhang, Jingfeng and Pan, Yulin and Liu, Yu},
  booktitle={Proceedings of the IEEE/CVF International Conference on Computer Vision},
  pages={17191--17202},
  year={2025}
}

@article{hu2023gaia,
  title={Gaia-1: A generative world model for autonomous driving},
  author={Hu, Anthony and Russell, Lloyd and Yeo, Hudson and Murez, Zak and Fedoseev, George and Kendall, Alex and Shotton, Jamie and Corrado, Gianluca},
  journal={arXiv preprint arXiv:2309.17080},
  year={2023}
}

@article{wang2026hand2world,
  title={Hand2world: Autoregressive egocentric interaction generation via free-space hand gestures},
  author={Wang, Yuxi and Ouyang, Wenqi and Wei, Tianyi and Dong, Yi and Shen, Zhiqi and Pan, Xingang},
  journal={arXiv preprint arXiv:2602.09600},
  year={2026}
}

@article{hao2026egosim,
  title={EgoSim: Egocentric World Simulator for Embodied Interaction Generation},
  author={Hao, Jinkun and Jia, Mingda and Wang, Ruiyan and Liu, Xihui and Yi, Ran and Ma, Lizhuang and Pang, Jiangmiao and Xu, Xudong},
  journal={arXiv preprint arXiv:2604.01001},
  year={2026}
}

@inproceedings{li2026mask2iv,
  title={Mask2iv: Interaction-centric video generation via mask trajectories},
  author={Li, Gen and Zhao, Bo and Yang, Jianfei and Sevilla-Lara, Laura},
  booktitle={Proceedings of the AAAI Conference on Artificial Intelligence},
  volume={40},
  pages={6091--6099},
  year={2026}
}

@article{zhang2025vhoi,
  title={VHOI: Controllable Video Generation of Human-Object Interactions from Sparse Trajectories via Motion Densification},
  author={Zhang, Wanyue and Foo, Lin Geng and Beeler, Thabo and Dabral, Rishabh and Theobalt, Christian},
  journal={arXiv preprint arXiv:2512.09646},
  year={2025}
}

@inproceedings{akkerman2025interdyn,
  title={Interdyn: Controllable interactive dynamics with video diffusion models},
  author={Akkerman, Rick and Feng, Haiwen and Black, Michael J and Tzionas, Dimitrios and Abrevaya, Victoria Fern{\'a}ndez},
  booktitle={Proceedings of the IEEE/CVF Conference on Computer Vision and Pattern Recognition},
  pages={12467--12479},
  year={2025}
}

@article{li2024zerohsi,
  title={Zerohsi: Zero-shot 4d human-scene interaction by video generation},
  author={Li, Hongjie and Yu, Hong-Xing and Li, Jiaman and Wu, Jiajun},
  journal={arXiv preprint arXiv:2412.18600},
  year={2024}
}

@inproceedings{wang2024motionctrl,
  title={Motionctrl: A unified and flexible motion controller for video generation},
  author={Wang, Zhouxia and Yuan, Ziyang and Wang, Xintao and Li, Yaowei and Chen, Tianshui and Xia, Menghan and Luo, Ping and Shan, Ying},
  booktitle={ACM SIGGRAPH 2024 Conference Papers},
  pages={1--11},
  year={2024}
}

@article{hong2022cogvideo,
  title={Cogvideo: Large-scale pretraining for text-to-video generation via transformers},
  author={Hong, Wenyi and Ding, Ming and Zheng, Wendi and Liu, Xinghan and Tang, Jie},
  journal={arXiv preprint arXiv:2205.15868},
  year={2022}
}

@misc{genrobot2025ego,
  author       = {{GenRobot}},
  title        = {{Gen-EgoData}},
  year         = {2025},
  howpublished = {Hugging Face Datasets},
  note         = {\url{https://huggingface.co/datasets/genrobot2025/Gen-EgoData}}
}

@misc{genrobot2025realomni,
  title        = {10Kh-RealOmni-OpenData},
  author       = {{GenRobot}},
  year         = {2025},
  howpublished = {\url{https://huggingface.co/datasets/genrobot2025/10Kh-RealOmin-OpenData}},
  note         = {Accessed: 2026-04-28}
}

@article{wan2025wan,
  title={Wan: Open and advanced large-scale video generative models},
  author={Wan, Team and Wang, Ang and Ai, Baole and Wen, Bin and Mao, Chaojie and Xie, Chen-Wei and Chen, Di and Yu, Feiwu and Zhao, Haiming and Yang, Jianxiao and others},
  journal={arXiv preprint arXiv:2503.20314},
  year={2025}
}

@inproceedings{li2026genhsi,
  title={Genhsi: Controllable generation of human-scene interaction videos},
  author={Li, Zekun and Zhou, Rui and Sajnani, Rahul and Cong, Xiaoyan and Ritchie, Daniel and Sridhar, Srinath},
  booktitle={Proceedings of the IEEE/CVF Winter Conference on Applications of Computer Vision},
  pages={138--149},
  year={2026}
}

@article{HOIGen,
  title={HOIGen-1M: A Large-scale Dataset for Human-Object Interaction Video Generation},
  author={Kun Liu and Qi Liu and Xinchen Liu and Jie Li and Yongdong Zhang and Jiebo Luo and Xiaodong He and Wu Liu},
  journal={arXiv preprint arXiv:2503.23715},
  year={2025}
}

@inproceedings{ge2018hand,
  title={Hand pointnet: 3d hand pose estimation using point sets},
  author={Ge, Liuhao and Cai, Yujun and Weng, Junwu and Yuan, Junsong},
  booktitle={Proceedings of the IEEE conference on computer vision and pattern recognition},
  pages={8417--8426},
  year={2018}
}

@inproceedings{xiao2024spatialtracker,
  title={Spatialtracker: Tracking any 2d pixels in 3d space},
  author={Xiao, Yuxi and Wang, Qianqian and Zhang, Shangzhan and Xue, Nan and Peng, Sida and Shen, Yujun and Zhou, Xiaowei},
  booktitle={Proceedings of the IEEE/CVF Conference on Computer Vision and Pattern Recognition},
  pages={20406--20417},
  year={2024}
}

@inproceedings{hu2025depthcrafter,
  title={Depthcrafter: Generating consistent long depth sequences for open-world videos},
  author={Hu, Wenbo and Gao, Xiangjun and Li, Xiaoyu and Zhao, Sijie and Cun, Xiaodong and Zhang, Yong and Quan, Long and Shan, Ying},
  booktitle={Proceedings of the IEEE/CVF Conference on Computer Vision and Pattern Recognition},
  pages={2005--2015},
  year={2025}
}

@article{guo2023animatediff,
  title={Animatediff: Animate your personalized text-to-image diffusion models without specific tuning},
  author={Guo, Yuwei and Yang, Ceyuan and Rao, Anyi and Liang, Zhengyang and Wang, Yaohui and Qiao, Yu and Agrawala, Maneesh and Lin, Dahua and Dai, Bo},
  journal={arXiv preprint arXiv:2307.04725},
  year={2023}
}

@article{xing2024make,
  title={Make-your-video: Customized video generation using textual and structural guidance},
  author={Xing, Jinbo and Xia, Menghan and Liu, Yuxin and Zhang, Yuechen and Zhang, Yong and He, Yingqing and Liu, Hanyuan and Chen, Haoxin and Cun, Xiaodong and Wang, Xintao and others},
  journal={IEEE transactions on visualization and computer graphics},
  volume={31},
  number={2},
  pages={1526--1541},
  year={2024},
  publisher={IEEE}
}

@inproceedings{xing2024dynamicrafter,
  title={Dynamicrafter: Animating open-domain images with video diffusion priors},
  author={Xing, Jinbo and Xia, Menghan and Zhang, Yong and Chen, Haoxin and Yu, Wangbo and Liu, Hanyuan and Liu, Gongye and Wang, Xintao and Shan, Ying and Wong, Tien-Tsin},
  booktitle={European Conference on Computer Vision},
  pages={399--417},
  year={2024},
  organization={Springer}
}

@inproceedings{zhao2024motiondirector,
  title={Motiondirector: Motion customization of text-to-video diffusion models},
  author={Zhao, Rui and Gu, Yuchao and Wu, Jay Zhangjie and Zhang, David Junhao and Liu, Jia-Wei and Wu, Weijia and Keppo, Jussi and Shou, Mike Zheng},
  booktitle={European Conference on Computer Vision},
  pages={273--290},
  year={2024},
  organization={Springer}
}

@inproceedings{jeong2024dreammotion,
  title={Dreammotion: Space-time self-similar score distillation for zero-shot video editing},
  author={Jeong, Hyeonho and Chang, Jinho and Park, Geon Yeong and Ye, Jong Chul},
  booktitle={European Conference on Computer Vision},
  pages={358--376},
  year={2024},
  organization={Springer}
}

@article{hu2024motionmaster,
  title={Motionmaster: Training-free camera motion transfer for video generation},
  author={Hu, Teng and Zhang, Jiangning and Yi, Ran and Wang, Yating and Huang, Hongrui and Weng, Jieyu and Wang, Yabiao and Ma, Lizhuang},
  journal={arXiv preprint arXiv:2404.15789},
  year={2024}
}

@article{yesiltepe2024motionshop,
  title={Motionshop: Zero-shot motion transfer in video diffusion models with mixture of score guidance},
  author={Yesiltepe, Hidir and Meral, Tuna Han Salih and Dunlop, Connor and Yanardag, Pinar},
  journal={arXiv preprint arXiv:2412.05355},
  year={2024}
}

@inproceedings{peebles2023scalable,
  title={Scalable diffusion models with transformers},
  author={Peebles, William and Xie, Saining},
  booktitle={Proceedings of the IEEE/CVF international conference on computer vision},
  pages={4195--4205},
  year={2023}
}

@inproceedings{pondaven2025video,
  title={Video motion transfer with diffusion transformers},
  author={Pondaven, Alexander and Siarohin, Aliaksandr and Tulyakov, Sergey and Torr, Philip and Pizzati, Fabio},
  booktitle={Proceedings of the Computer Vision and Pattern Recognition Conference},
  pages={22911--22921},
  year={2025}
}

@article{huang2025vipe,
  title={Vipe: Video pose engine for 3d geometric perception},
  author={Huang, Jiahui and Zhou, Qunjie and Rabeti, Hesam and Korovko, Aleksandr and Ling, Huan and Ren, Xuanchi and Shen, Tianchang and Gao, Jun and Slepichev, Dmitry and Lin, Chen-Hsuan and others},
  journal={arXiv preprint arXiv:2508.10934},
  year={2025}
}

@article{zhao2023controlvideo,
  title={Controlvideo: Adding conditional control for one shot text-to-video editing},
  author={Zhao, Min and Wang, Rongzhen and Bao, Fan and Li, Chongxuan and Zhu, Jun},
  journal={arXiv preprint arXiv:2305.17098},
  volume={2},
  number={3},
  year={2023}
}

@article{kong2024hunyuanvideo,
  title={Hunyuanvideo: A systematic framework for large video generative models},
  author={Kong, Weijie and Tian, Qi and Zhang, Zijian and Min, Rox and Dai, Zuozhuo and Zhou, Jin and Xiong, Jiangfeng and Li, Xin and Wu, Bo and Zhang, Jianwei and others},
  journal={arXiv preprint arXiv:2412.03603},
  year={2024}
}

@inproceedings{sudhakar2024controlling,
  title={Controlling the world by sleight of hand},
  author={Sudhakar, Sruthi and Liu, Ruoshi and Hoorick, Basile Van and Vondrick, Carl and Zemel, Richard},
  booktitle={European Conference on Computer Vision},
  pages={414--430},
  year={2024},
  organization={Springer}
}

@inproceedings{zhang2025framepainter,
  title={Framepainter: Endowing interactive image editing with video diffusion priors},
  author={Zhang, Yabo and Zhou, Xinpeng and Zeng, Yihan and Xu, Hang and Li, Hui and Zuo, Wangmeng},
  booktitle={Proceedings of the IEEE/CVF International Conference on Computer Vision},
  pages={18121--18131},
  year={2025}
}

@inproceedings{ren2024customize,
  title={Customize-a-video: One-shot motion customization of text-to-video diffusion models},
  author={Ren, Yixuan and Zhou, Yang and Yang, Jimei and Shi, Jing and Liu, Difan and Liu, Feng and Kwon, Mingi and Shrivastava, Abhinav},
  booktitle={European Conference on Computer Vision},
  pages={332--349},
  year={2024},
  organization={Springer}
}

@article{wu2024motionbooth,
  title={Motionbooth: Motion-aware customized text-to-video generation},
  author={Wu, Jianzong and Li, Xiangtai and Zeng, Yanhong and Zhang, Jiangning and Zhou, Qianyu and Li, Yining and Chen, Kai and Tong, Yunhai},
  journal={Advances in Neural Information Processing Systems},
  volume={37},
  pages={34322--34348},
  year={2024}
}

@article{geyer2023tokenflow,
  title={Tokenflow: Consistent diffusion features for consistent video editing},
  author={Geyer, Michal and Bar-Tal, Omer and Bagon, Shai and Dekel, Tali},
  journal={arXiv preprint arXiv:2307.10373},
  year={2023}
}

@inproceedings{qi2023fatezero,
  title={Fatezero: Fusing attentions for zero-shot text-based video editing},
  author={Qi, Chenyang and Cun, Xiaodong and Zhang, Yong and Lei, Chenyang and Wang, Xintao and Shan, Ying and Chen, Qifeng},
  booktitle={Proceedings of the IEEE/CVF International Conference on Computer Vision},
  pages={15932--15942},
  year={2023}
}

@article{ling2024motionclone,
  title={Motionclone: Training-free motion cloning for controllable video generation},
  author={Ling, Pengyang and Bu, Jiazi and Zhang, Pan and Dong, Xiaoyi and Zang, Yuhang and Wu, Tong and Chen, Huaian and Wang, Jiaqi and Jin, Yi},
  journal={arXiv preprint arXiv:2406.05338},
  year={2024}
}

@article{xiao2024video,
  title={Video diffusion models are training-free motion interpreter and controller},
  author={Xiao, Zeqi and Zhou, Yifan and Yang, Shuai and Pan, Xingang},
  journal={Advances in Neural Information Processing Systems},
  volume={37},
  pages={76115--76138},
  year={2024}
}

@inproceedings{abdal2025dynamic,
  title={Dynamic concepts personalization from single videos},
  author={Abdal, Rameen and Patashnik, Or and Skorokhodov, Ivan and Menapace, Willi and Siarohin, Aliaksandr and Tulyakov, Sergey and Cohen-Or, Daniel and Aberman, Kfir},
  booktitle={Proceedings of the Special Interest Group on Computer Graphics and Interactive Techniques Conference Conference Papers},
  pages={1--9},
  year={2025}
}

@inproceedings{shi2025decouple,
  title={Decouple and track: Benchmarking and improving video diffusion transformers for motion transfer},
  author={Shi, Qingyu and Wu, Jianzong and Bai, Jinbin and Zhang, Jiangning and Qi, Lu and Tong, Yunhai and Li, Xiangtai},
  booktitle={Proceedings of the IEEE/CVF International Conference on Computer Vision},
  pages={10995--11005},
  year={2025}
}

@article{ma2026fastvmt,
  title={FastVMT: Eliminating Redundancy in Video Motion Transfer},
  author={Ma, Yue and Wang, Zhikai and Ren, Tianhao and Zheng, Mingzhe and Liu, Hongyu and Guo, Jiayi and Fong, Mark and Xue, Yuxuan and Zhao, Zixiang and Schindler, Konrad and others},
  journal={arXiv preprint arXiv:2602.05551},
  year={2026}
}

@article{liang2025comprehensive,
  title={A comprehensive survey on test-time adaptation under distribution shifts},
  author={Liang, Jian and He, Ran and Tan, Tieniu},
  journal={International Journal of Computer Vision},
  volume={133},
  number={1},
  pages={31--64},
  year={2025},
  publisher={Springer}
}

@inproceedings{sun2020test,
  title={Test-time training with self-supervision for generalization under distribution shifts},
  author={Sun, Yu and Wang, Xiaolong and Liu, Zhuang and Miller, John and Efros, Alexei and Hardt, Moritz},
  booktitle={International conference on machine learning},
  pages={9229--9248},
  year={2020},
  organization={PMLR}
}

@article{wang2025test,
  title={Test-time training on video streams},
  author={Wang, Renhao and Sun, Yu and Tandon, Arnuv and Gandelsman, Yossi and Chen, Xinlei and Efros, Alexei A and Wang, Xiaolong},
  journal={Journal of Machine Learning Research},
  volume={26},
  number={9},
  pages={1--29},
  year={2025}
}

@inproceedings{liu2025video,
  title={Video-t1: Test-time scaling for video generation},
  author={Liu, Fangfu and Wang, Hanyang and Cai, Yimo and Zhang, Kaiyan and Zhan, Xiaohang and Duan, Yueqi},
  booktitle={Proceedings of the IEEE/CVF international conference on computer vision},
  pages={18671--18681},
  year={2025}
}

@article{he2025scaling,
  title={Scaling image and video generation via test-time evolutionary search},
  author={He, Haoran and Liang, Jiajun and Wang, Xintao and Wan, Pengfei and Zhang, Di and Gai, Kun and Pan, Ling},
  journal={arXiv preprint arXiv:2505.17618},
  year={2025}
}

@inproceedings{wallace2024diffusion,
  title={Diffusion model alignment using direct preference optimization},
  author={Wallace, Bram and Dang, Meihua and Rafailov, Rafael and Zhou, Linqi and Lou, Aaron and Purushwalkam, Senthil and Ermon, Stefano and Xiong, Caiming and Joty, Shafiq and Naik, Nikhil},
  booktitle={Proceedings of the IEEE/CVF Conference on Computer Vision and Pattern Recognition},
  pages={8228--8238},
  year={2024}
}

@article{tong2025delving,
  title={Delving into rl for image generation with cot: A study on dpo vs. grpo},
  author={Tong, Chengzhuo and Guo, Ziyu and Zhang, Renrui and Shan, Wenyu and Wei, Xinyu and Xing, Zhenghao and Li, Hongsheng and Heng, Pheng-Ann},
  journal={arXiv preprint arXiv:2505.17017},
  year={2025}
}

@article{xie2025sana,
  title={Sana 1.5: Efficient scaling of training-time and inference-time compute in linear diffusion transformer},
  author={Xie, Enze and Chen, Junsong and Zhao, Yuyang and Yu, Jincheng and Zhu, Ligeng and Wu, Chengyue and Lin, Yujun and Zhang, Zhekai and Li, Muyang and Chen, Junyu and others},
  journal={arXiv preprint arXiv:2501.18427},
  year={2025}
}

@inproceedings{zhuo2025reflection,
  title={From reflection to perfection: Scaling inference-time optimization for text-to-image diffusion models via reflection tuning},
  author={Zhuo, Le and Zhao, Liangbing and Paul, Sayak and Liao, Yue and Zhang, Renrui and Xin, Yi and Gao, Peng and Elhoseiny, Mohamed and Li, Hongsheng},
  booktitle={Proceedings of the IEEE/CVF International Conference on Computer Vision},
  pages={15329--15339},
  year={2025}
}

@article{ramesh2025test,
  title={Test-time scaling of diffusion models via noise trajectory search},
  author={Ramesh, Vignav and Mardani, Morteza},
  journal={arXiv preprint arXiv:2506.03164},
  year={2025}
}

@article{singhal2025general,
  title={A general framework for inference-time scaling and steering of diffusion models},
  author={Singhal, Raghav and Horvitz, Zachary and Teehan, Ryan and Ren, Mengye and Yu, Zhou and McKeown, Kathleen and Ranganath, Rajesh},
  journal={arXiv preprint arXiv:2501.06848},
  year={2025}
}

@article{schneider2020improving,
  title={Improving robustness against common corruptions by covariate shift adaptation},
  author={Schneider, Steffen and Rusak, Evgenia and Eck, Luisa and Bringmann, Oliver and Brendel, Wieland and Bethge, Matthias},
  journal={Advances in neural information processing systems},
  volume={33},
  pages={11539--11551},
  year={2020}
}

@inproceedings{liang2020we,
  title={Do we really need to access the source data? source hypothesis transfer for unsupervised domain adaptation},
  author={Liang, Jian and Hu, Dapeng and Feng, Jiashi},
  booktitle={International conference on machine learning},
  pages={6028--6039},
  year={2020},
  organization={PMLR}
}

@inproceedings{shu2021open,
  title={Open domain generalization with domain-augmented meta-learning},
  author={Shu, Yang and Cao, Zhangjie and Wang, Chenyu and Wang, Jianmin and Long, Mingsheng},
  booktitle={Proceedings of the IEEE/CVF conference on computer vision and pattern recognition},
  pages={9624--9633},
  year={2021}
}

@article{zhang2022memo,
  title={Memo: Test time robustness via adaptation and augmentation},
  author={Zhang, Marvin and Levine, Sergey and Finn, Chelsea},
  journal={Advances in neural information processing systems},
  volume={35},
  pages={38629--38642},
  year={2022}
}

@article{gandelsman2022test,
  title={Test-time training with masked autoencoders},
  author={Gandelsman, Yossi and Sun, Yu and Chen, Xinlei and Efros, Alexei},
  journal={Advances in Neural Information Processing Systems},
  volume={35},
  pages={29374--29385},
  year={2022}
}

@article{eyring2025noise,
  title={Noise hypernetworks: Amortizing test-time compute in diffusion models},
  author={Eyring, Luca and Karthik, Shyamgopal and Dosovitskiy, Alexey and Ruiz, Nataniel and Akata, Zeynep},
  journal={arXiv preprint arXiv:2508.09968},
  year={2025}
}

@article{yu2025autorefiner,
  title={AutoRefiner: Improving Autoregressive Video Diffusion Models via Reflective Refinement Over the Stochastic Sampling Path},
  author={Yu, Zhengyang and Hayakawa, Akio and Ishii, Masato and Yu, Qingtao and Shibuya, Takashi and Zhang, Jing and Mitsufuji, Yuki},
  journal={arXiv preprint arXiv:2512.11203},
  year={2025}
}

@article{hong2024slowfast,
  title={Slowfast-vgen: Slow-fast learning for action-driven long video generation},
  author={Hong, Yining and Liu, Beide and Wu, Maxine and Zhai, Yuanhao and Chang, Kai-Wei and Li, Linjie and Lin, Kevin and Lin, Chung-Ching and Wang, Jianfeng and Yang, Zhengyuan and others},
  journal={arXiv preprint arXiv:2410.23277},
  year={2024}
}

@article{kim2025test,
  title={Test-time alignment of diffusion models without reward over-optimization},
  author={Kim, Sunwoo and Kim, Minkyu and Park, Dongmin},
  journal={arXiv preprint arXiv:2501.05803},
  year={2025}
}

@article{shu2022test,
  title={Test-time prompt tuning for zero-shot generalization in vision-language models},
  author={Shu, Manli and Nie, Weili and Huang, De-An and Yu, Zhiding and Goldstein, Tom and Anandkumar, Anima and Xiao, Chaowei},
  journal={Advances in Neural Information Processing Systems},
  volume={35},
  pages={14274--14289},
  year={2022}
}

@article{li2025test,
  title={Test-time preference optimization: On-the-fly alignment via iterative textual feedback},
  author={Li, Yafu and Hu, Xuyang and Qu, Xiaoye and Li, Linjie and Cheng, Yu},
  journal={arXiv preprint arXiv:2501.12895},
  year={2025}
}

@article{lian2023llm,
  title={Llm-grounded video diffusion models},
  author={Lian, Long and Shi, Baifeng and Yala, Adam and Darrell, Trevor and Li, Boyi},
  journal={arXiv preprint arXiv:2309.17444},
  year={2023}
}

@inproceedings{wu2024self,
  title={Self-correcting llm-controlled diffusion models},
  author={Wu, Tsung-Han and Lian, Long and Gonzalez, Joseph E and Li, Boyi and Darrell, Trevor},
  booktitle={Proceedings of the IEEE/CVF Conference on Computer Vision and Pattern Recognition},
  pages={6327--6336},
  year={2024}
}

@inproceedings{blattmann2023align,
  title={Align your latents: High-resolution video synthesis with latent diffusion models},
  author={Blattmann, Andreas and Rombach, Robin and Ling, Huan and Dockhorn, Tim and Kim, Seung Wook and Fidler, Sanja and Kreis, Karsten},
  booktitle={Proceedings of the IEEE/CVF conference on computer vision and pattern recognition},
  pages={22563--22575},
  year={2023}
}

@inproceedings{ma2024followpose,
  title={Follow your pose: Pose-guided text-to-video generation using pose-free videos},
  author={Ma, Yue and He, Yingqing and Cun, Xiaodong and Wang, Xintao and Chen, Siran and Li, Xiu and Chen, Qifeng},
  booktitle={Proceedings of the AAAI Conference on Artificial Intelligence},
  volume={38},
  number={5},
  pages={4117--4125},
  year={2024}
}

@article{ma2025followyourmotion,
  title={Follow-Your-Motion: Video Motion Transfer via Efficient Spatial-Temporal Decoupled Finetuning},
  author={Ma, Yue and Liu, Yulong and Zhu, Qiyuan and Yang, Ayden and Feng, Kunyu and Zhang, Xinhua and Li, Zhifeng and Han, Sirui and Qi, Chenyang and Chen, Qifeng},
  journal={arXiv preprint arXiv:2506.05207},
  year={2025}
}

@article{wang2024cove,
  title={Cove: Unleashing the diffusion feature correspondence for consistent video editing},
  author={Wang, Jiangshan and Ma, Yue and Guo, Jiayi and Xiao, Yicheng and Huang, Gao and Li, Xiu},
  journal={Advances in Neural Information Processing Systems},
  volume={37},
  pages={96541--96565},
  year={2024}
}

@article{wang2026liveedit,
  title={LiveEdit: Towards Real-Time Diffusion-Based Streaming Video Editing},
  author={Wang, Xinyu and Zhao, Chongbo and Zhan, Fangneng and Ma, Yue},
  journal={arXiv preprint arXiv:2606.26740},
  year={2026}
}

@article{liu2026opsd,
  title={OPSD-V: On-Policy Self-Distillation for Post-Training Few-Step Autoregressive Video Generators},
  author={Liu, Hongyu and Wang, Chun and Gao, Feng and He, Xuanhua and Ma, Yue and Wan, Ziyu and Zhang, Yong and Wei, Xiaoming and Chen, Qifeng},
  journal={arXiv preprint arXiv:2607.08766},
  year={2026}
}

@article{wang2024taming,
  title={Taming rectified flow for inversion and editing},
  author={Wang, Jiangshan and Pu, Junfu and Qi, Zhongang and Guo, Jiayi and Ma, Yue and Huang, Nisha and Chen, Yuxin and Li, Xiu and Shan, Ying},
  journal={arXiv preprint arXiv:2411.04746},
  year={2024}
}

@inproceedings{feng2025dit4edit,
  title={Dit4edit: Diffusion transformer for image editing},
  author={Feng, Kunyu and Ma, Yue and Wang, Bingyuan and Qi, Chenyang and Chen, Haozhe and Chen, Qifeng and Wang, Zeyu},
  booktitle={Proceedings of the AAAI Conference on Artificial Intelligence},
  volume={39},
  number={3},
  pages={2969--2977},
  year={2025}
}

@inproceedings{qiu2024tfb,
title   = {{TFB}: Towards Comprehensive and Fair Benchmarking of Time Series Forecasting Methods},
author  = {Xiangfei Qiu and Jilin Hu and Lekui Zhou and Xingjian Wu and Junyang Du and Buang Zhang and Chenjuan Guo and Aoying Zhou and Christian S. Jensen and Zhenli Sheng and Bin Yang},
booktitle = {Proc. {VLDB} Endow.},
pages   = {2363--2377},
year    = {2024}
}

@inproceedings{qiu2025duet,
title   = {{DUET}: Dual Clustering Enhanced Multivariate Time Series Forecasting},
author  = {Xiangfei Qiu and Xingjian Wu and Yan Lin and Chenjuan Guo and Jilin Hu and Bin Yang},
booktitle = {SIGKDD},
pages     = {1185-1196},
year    = {2025}
}

@inproceedings{qiu2025DBLoss,
title   = {DBLoss: Decomposition-based Loss Function for Time Series Forecasting},
author  = {Xiangfei Qiu and Xingjian Wu and Hanyin Cheng and Xvyuan Liu and Chenjuan Guo and Jilin Hu and Bin Yang},
booktitle = {NeurIPS},
year    = {2025}
}

@inproceedings{qiu2026dag,
  title={DAG: A Dual Correlation Network for Time Series Forecasting with Exogenous Variables},
  author={Qiu, Xiangfei and Zhu, Yuhan and Li, Zhengyu and Wu, Xingjian and Yang, Bin and Hu, Jilin},
  booktitle={ICML},
  year={2026}
}

@article{zhang2026kvpo,
  title={Kvpo: Ode-native grpo for autoregressive video alignment via kv semantic exploration},
  author={Zhang, Ruicheng and Cong, Kaixi and Zhou, Jun and Zhong, Zhizhou and Xu, Zunnan and Mao, Shuiyang and Liu, Wei and Li, Xiu},
  journal={arXiv preprint arXiv:2605.14278},
  year={2026}
}

@inproceedings{zhang2026zo3t,
  title={Zo3t: Zero-shot 3d-aware trajectory-guided image-to-video generation via test-time training},
  author={Zhang, Ruicheng and Zhou, Jun and Xu, Zunnan and Liu, Zihao and Huang, Jiehui and Zhang, Mingyang and Sun, Yu and Li, Xiu},
  booktitle={Proceedings of the AAAI Conference on Artificial Intelligence},
  volume={40},
  number={15},
  pages={12708--12716},
  year={2026}
}

@inproceedings{chen2026taming,
  title={Taming preference mode collapse via directional decoupling alignment in diffusion reinforcement learning},
  author={Chen, Chubin and Hu, Sujie and Zhu, Jiashu and Wu, Meiqi and Chen, Jintao and Li, Yanxun and Huang, Nisha and Fang, Chengyu and Wu, Jiahong and Chu, Xiangxiang and others},
  booktitle={Proceedings of the IEEE/CVF Conference on Computer Vision and Pattern Recognition},
  pages={12775--12786},
  year={2026}
}

@article{hu2026embedding,
  title={Embedding-perturbed exploration preference optimization for flow models},
  author={Hu, Sujie and Chen, Chubin and Zhu, Jiashu and Wu, Jiahong and Chu, Xiangxiang and Li, Xiu},
  journal={arXiv preprint arXiv:2605.15803},
  year={2026}
}

@article{xu2026disco,
  title={Disco-LoRA: Disentangled Composition of Content, Style, and Motion for Multi-concept Video Customization},
  author={Xu, Xuancheng and Jia, Gengyun and Bao, Bing-Kun},
  journal={arXiv preprint arXiv:2606.26668},
  year={2026}
}

@inproceedings{xu2025clgc,
  title={Clgc: Continuous layout guidance for consistent text-to-video editing},
  author={Xu, Xuancheng and Tao, Ming and Bao, Bing-Kun},
  booktitle={2025 IEEE International Conference on Multimedia and Expo (ICME)},
  pages={1--6},
  year={2025},
  organization={IEEE}
}

@article{song2026vista,
  title={VISTA: Triplet-Supervised Video Style Transfer with Diffusion Transformers},
  author={Song, Yiren and Yao, Wangzi and Wang, Haofan and Shou, Mike Zheng},
  journal={arXiv preprint arXiv:2605.17312},
  year={2026}
}

@article{song2026streamingeffect,
  title={StreamingEffect: Real-Time Human-Centric Video Effect Generation},
  author={Song, Yiren and Liu, Cheng and Jiang, Yuxin and Shou, Mike Zheng},
  journal={arXiv preprint arXiv:2605.17019},
  year={2026}
}

@article{gao2026pai,
  title={PAI-Studio: Cinematic Video Background Replacement with Camera-Aware Motion},
  author={Gao, Heyuan and Tang, Bangxun and Song, Yiren and Fang, Guian and He, Zijian and Yang, Jie and Shou, Mike Zheng},
  journal={arXiv preprint arXiv:2606.01399},
  year={2026}
}

@inproceedings{song2024processpainter,
  title={ProcessPainter: Learning to draw from sequence data},
  author={Song, Yiren and Huang, Shijie and Yao, Chen and Ci, Hai and Ye, Xiaojun and Liu, Jiaming and Zhang, Yuxuan and Shou, Mike Zheng},
  booktitle={SIGGRAPH Asia 2024 Conference Papers},
  pages={1--10},
  year={2024}
}
}

\clearpage

\appendix

\section{Related Work}

\noindent\textbf{Test-Time Visual Generation.}
Video generation~\cite{ma2024followpose,  song2026streamingeffect,ma2025followyourmotion, wang2024cove, wang2026liveedit, liu2026opsd, wang2024taming} has advanced rapidly, largely driven by recent progress in diffusion models. Test-time methods enhance pre-trained models during inference without retraining~\cite{liang2025comprehensive, sun2020test, wang2025test}. Existing approaches primarily follow two paradigms. First, test-time scaling improves quality through iterative candidate search~\cite{liu2025video, he2025scaling}, preference optimization~\cite{wallace2024diffusion, tong2025delving}, or VLM-guided reflection~\cite{xie2025sana, zhuo2025reflection}. However, this process requires prohibitive computational costs and shows marginal improvements for continuous diffusion models due to inherent architectural constraints~\cite{ramesh2025test, singhal2025general}. Second, test-time optimization refines outputs via auxiliary parameter updates~\cite{schneider2020improving, liang2020we, shu2021open, zhang2022memo, zhang2026kvpo, zhang2026zo3t, chen2026taming, hu2026embedding, xu2026disco, song2026vista, gao2026pai, song2024processpainter, xu2025clgc,  gandelsman2022test, eyring2025noise, yu2025autorefiner, hong2024slowfast}, prompt tuning~\cite{shu2022test}, or layout-guided latent optimization~\cite{kim2025test, li2025test, lian2023llm, wu2024self}, but often relies on complex instance-specific training or global objectives. In contrast, our test-time refinements target the hand-object structure and local correspondences needed for egocentric action transfer.

\noindent\textbf{Video Diffusion Model.}

% \section{More details}
\section{More Implementation Details}
We use WAN-2.1~\cite{wan2025wan} as the base video generation model and keep all model parameters frozen during inference. For ablation study, all ablation variants are evaluated under the same setting as the main comparison, with only the specified component removed or modified. The latent video is initialized with Gaussian noise of shape $1\times16\times(\frac{F-1}{4}+1)\times\frac{H}{8}\times\frac{W}{8}$, where $F$ is internally rounded to $4k+1$. We enable latent tiling in the VAE space with \texttt{tile\_size}=$(30,52)$ and \texttt{tile\_stride}=$(15,26)$, which produces a per-frame DiT token grid of $h=\frac{H}{8}$ by $w=\frac{W}{8}$.

\paragraph{AMF-based structural mask.}
As described in the main text (Sec.~\ref{subsec:VSA}), AMF is used solely to construct the soft structural mask $\mathcal{M}$ for VSA, not as a separate optimization objective. For sliding-window AMF computation, we use a tile grid of $3\times4$ tokens, temporal span $s_f=5$, local search window $l=21$ with half-width 10, and temperature $\tau=1.0$. The AMF magnitude from the reference branch is computed over temporal offsets with linear weights decayed from 1.0 to 0.8 across offsets, normalized by the number of offsets and tokens, and then converted into the structural saliency mask via Eq.~\ref{eq:eq1}--\ref{eq:eq2}. The actual optimization objective is the mask-weighted velocity regression $\mathcal{L}_{\mathrm{VSA}}$ (Eq.~\ref{eq:eq3}), and gradients are applied only to the latent variable. To reduce computational cost during the inner optimization loop, we recompute Q/K features and tile-level AMF every 3 steps and reuse the cached mask on intermediate steps.

\paragraph{Optimization.}
At each guided diffusion step $t$, we form the reference latent by adding step-consistent noise to the cached clean latent and run a null-text forward pass to extract query/key features from the 15th DiT block. Each guided step performs 10 latent-only AdamW updates with a linear learning-rate decay from 0.003 to 0.002. The VSA weight, SCC correspondence weight, and stay regularization weight are set to 15.0, 1.0, and 0.05, respectively.

\paragraph{Inference time.} On an NVIDIA A100 GPU, the average inference time for a 81-frame video of resolution $832\times480$ is approximately 413 seconds, which includes the base diffusion process and the additional test-time optimization steps. The VSA module adds around 23 seconds, while the SCC module contributes an additional 72 seconds. The overall time is still within a practical range for test-time refinement, given the improved quality and consistency of the generated videos.

 \begin{figure*}
   \centering
   \includegraphics[width=\linewidth]{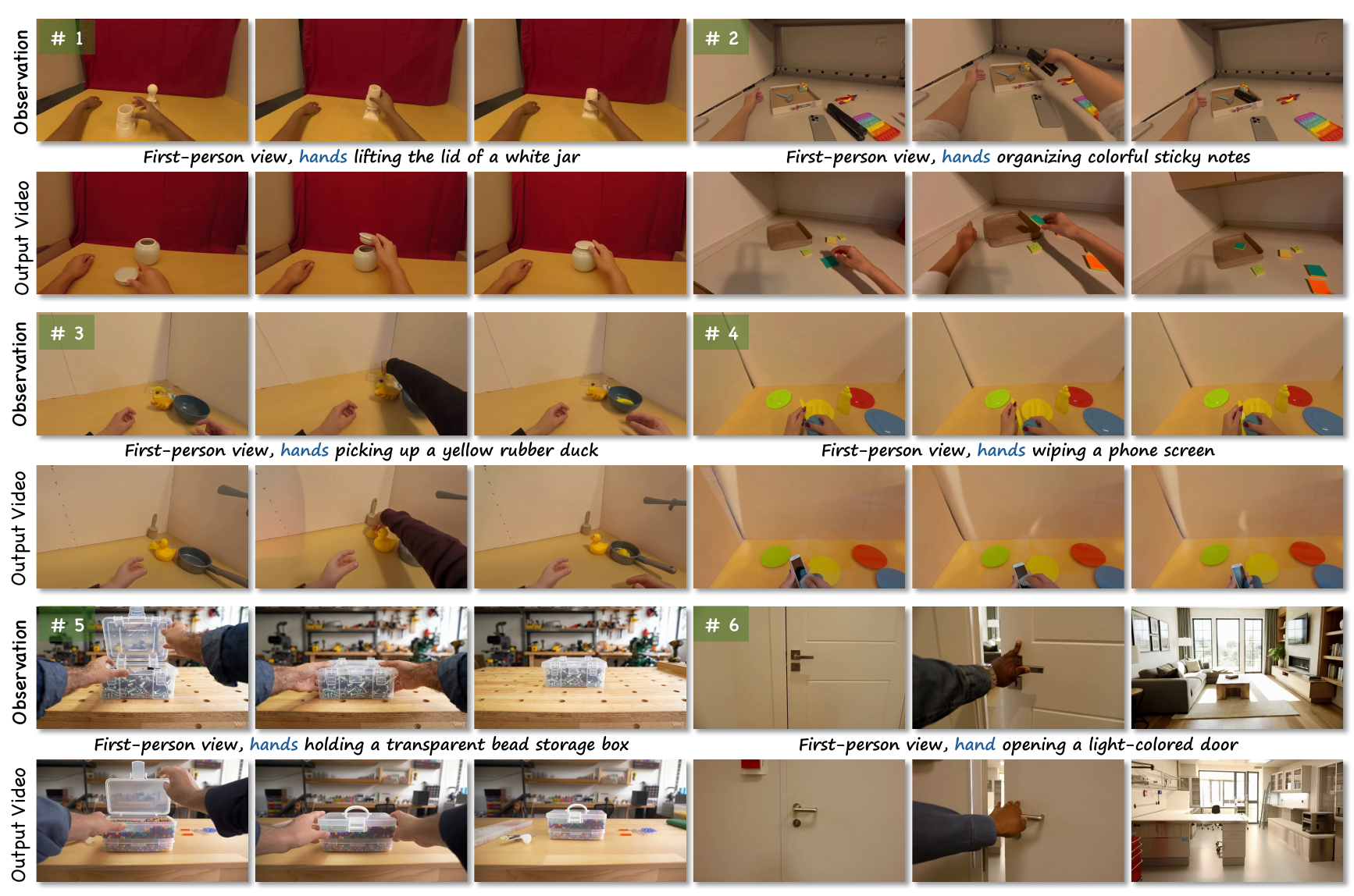}
   \caption{More visual results.}
   \label{fig:gallery_supple1}
 \end{figure*}

\section{More Details about Attention Motion Flow}
Attention Motion Flow (AMF)~\cite{ma2026fastvmt} extracts implicit motion cues from the self-attention layers of a video diffusion transformer. The key observation is that cross-frame query-key affinity in DiT attention naturally encodes patch-level correspondences: if a token in one frame attends strongly to a token in a neighboring frame, their spatial offset provides an estimate of the local motion. Instead of using external trackers or optical-flow estimators, we compute AMF directly from the internal attention features of the frozen generation model. To instantiate this intuition, we compute token-level correspondences from local query-key affinities within self-attention.

\begin{figure*}
   \centering
   \includegraphics[width=\linewidth]{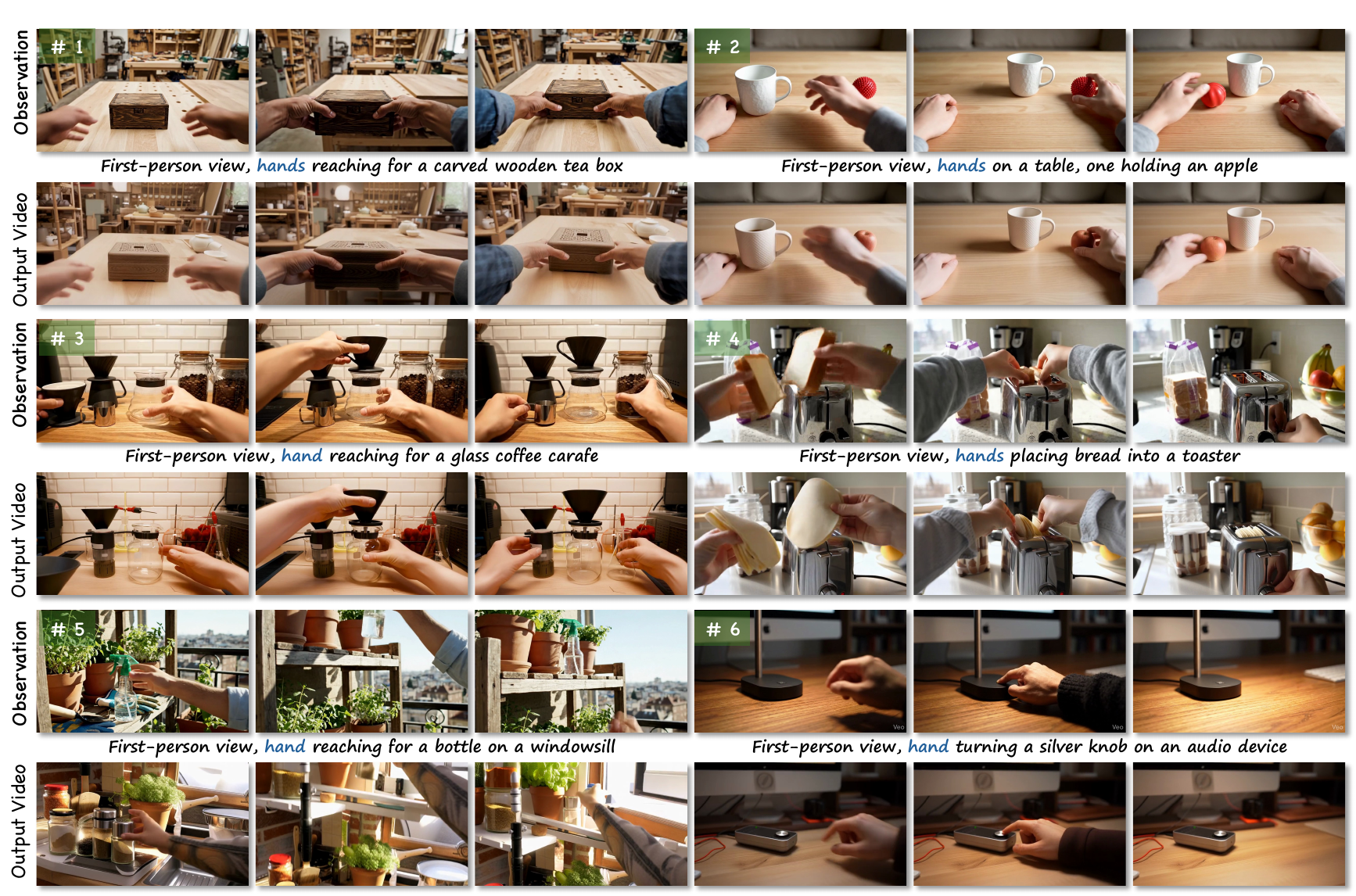}
   \caption{More visual results.}
   \label{fig:gallery_supple2}
 \end{figure*}

% \paragraph{Sliding-window Correspondence Extraction.}
\paragraph{Local Attention-based Correspondence.}
For a spatial token $u$ in frame $i$, we search its correspondence in a nearby frame $j$ within a local displacement window $\mathcal N(u)=\{u+\Delta\mid \Delta\in[-r,r]^2\}$. Let $q_i(u)$ be the query feature at token $u$ and $k_j(u+\Delta)$ be the key feature at the candidate location. We compute the local attention score as:
\begin{equation}
s_{i\rightarrow j}(u,\Delta)=\frac{\langle q_i(u),k_j(u+\Delta)\rangle}{\sqrt{d}},
\quad \Delta\in[-r,r]^2 ,
\end{equation}
where $d$ is the feature dimension. The normalized local correspondence distribution is then
\begin{equation}
p_{i\rightarrow j}(u,\Delta)=
\frac{\exp(s_{i\rightarrow j}(u,\Delta))}
{\sum_{\Delta'\in[-r,r]^2}\exp(s_{i\rightarrow j}(u,\Delta'))}.
\end{equation}
Following the soft motion extraction strategy in FastVMT, we use the expectation over local displacements to obtain a differentiable AMF vector:
\begin{equation}
\mathcal A_{i\rightarrow j}(u)=
\sum_{\Delta\in[-r,r]^2} p_{i\rightarrow j}(u,\Delta)\Delta .
\end{equation}
This sliding-window design reflects the local continuity of video motion and avoids unreliable global matching between unrelated tokens.

\begin{figure*}
   \centering
   \includegraphics[width=\linewidth]{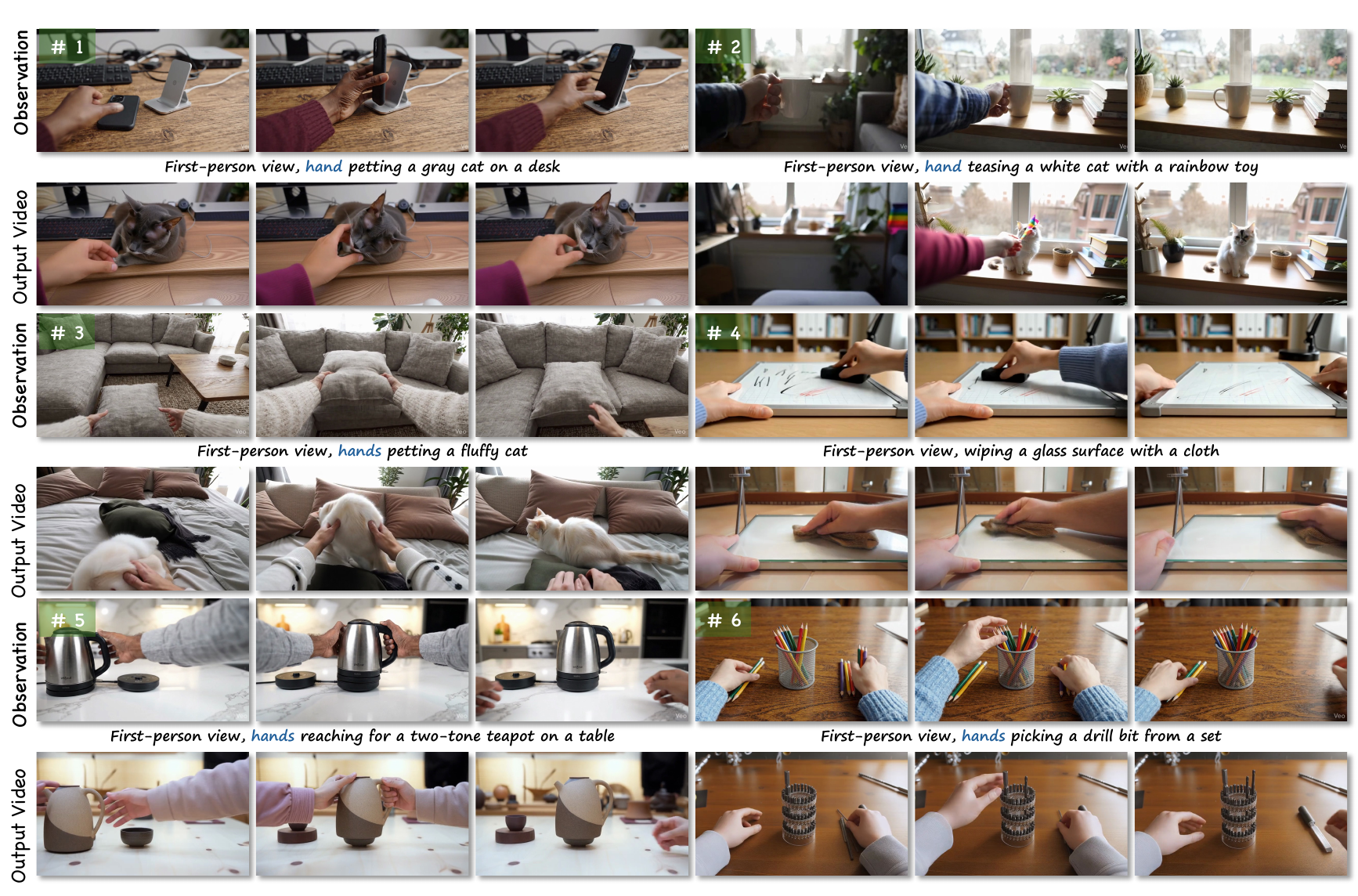}
   \caption{More visual results.}
   \label{fig:gallery_supple3}
 \end{figure*} 

\paragraph{AMF-induced Structural Mask.}
% In our VSA module, AMF is used to identify action-critical regions rather than as a direct motion constraint. We compute AMF from the reference branch and use its magnitude as a structural saliency score, i.e., $\mathcal S(u)=\|\mathcal A_{\mathrm{ref}}(u)\|_2$. Tokens with larger AMF magnitude usually correspond to moving hands, manipulated objects, or contact regions, and thus receive stronger velocity anchoring. Static regions are assigned a smaller but non-zero weight to preserve global scene coherence. After normalization and power reweighting, the AMF saliency produces the soft mask $\mathcal M$ used by VSA in Sec.~\ref{subsec:VSA}, enabling the optimization to focus on interaction-relevant structures while avoiding unnecessary constraints on background regions.

In our VSA module, AMF is not used as a hard motion supervision signal; instead, it serves as a cue for locating interaction-relevant structures. We compute AMF from the reference branch and convert its magnitude into a structural saliency score:

\begin{equation}
\mathcal S(u)=\|\mathcal A_{\mathrm{ref}}(u)\|_2,
\end{equation}

where larger values usually indicate regions with stronger action dynamics, such as moving hands, manipulated objects, and hand-object contact areas. Intuitively, these regions carry most of the structural and physical constraints of egocentric manipulation, so they should receive stronger anchoring during optimization.

After normalizing the structural saliency score over with Eq.~\ref{eq:eq2}, we then apply a power reweighting to control the concentration of the mask:

\begin{equation}
\mathcal M(u)=\mathcal {\hat S}(u)^\gamma (1-m_{\min})+m_{\min}.
\end{equation}

where,  $\gamma>1$ sharpens the mask and emphasizes the most motion-salient regions, while $\gamma<1$ produces a flatter weighting profile. The lower bound $m_{\min}$ ensures that low-motion regions still contribute weakly to the velocity anchoring objective, which helps preserve global scene layout and prevents the optimization from overfitting only to highly dynamic tokens.

Therefore, the AMF-induced mask allocates optimization strength adaptively: dynamic interaction regions are emphasized, whereas static regions are softly regularized. This design allows VSA in Sec.~\ref{subsec:VSA} to focus on the action-critical structure without imposing unnecessarily strong constraints on unrelated background content.

\section{More Details about User Study}
To complement automatic metrics, we conduct a user study with 20 volunteers on 50 randomly selected cases from EgoActionBench. For each evaluated case, volunteers are shown the reference egocentric manipulation video, the target text prompt, and generated videos from all compared methods. The method names are hidden, and the generated videos are presented in randomized order to reduce ordering bias.

% \paragraph{Visualization of the Questionnaire for User Study.} 

\paragraph{Rating Dimensions.}
Annotators evaluate each generated result according to the following four criteria:
\begin{list}{$\bullet$}{\leftmargin=1.2em \labelwidth=0.8em \labelsep=0.4em \itemsep=1pt \topsep=2pt}
    \item \textbf{Appearance Consistency:} whether the generated video preserves a coherent target appearance across frames without obvious identity drift, texture inconsistency, or visual artifacts.
    \item \textbf{Text Alignment:} whether the generated video follows the target prompt, including the described scene, object appearance, and intended manipulation context.
    \item \textbf{Temporal Consistency:} whether the generated video remains stable over time without obvious flickering, jittering, ghosting, or frame-to-frame structural instability.
    \item \textbf{Motion Fidelity:} whether the generated video preserves the reference action pattern, including hand motion, contact timing, and object response, without severe drifting or motion collapse.
\end{list}

\paragraph{Scoring Scale.}
Each dimension is rated using a four-level ordinal scale: \textbf{Excellent} (fully satisfies the criterion with no obvious artifacts), \textbf{Good} (minor artifacts or deviations), \textbf{Pass} (noticeable but acceptable issues), and \textbf{Fail} (severe artifacts or failure to satisfy the criterion). The final statistics are obtained by aggregating ratings across all volunteers and evaluated cases.

\section{More Details about EgoActionBench}

EgoActionBench focuses on whether the generated video preserves the interaction logic of the reference action, including contact transitions, object response, and coordinated hand-object dynamics.
The main paper summarizes the data sources, category coverage, and basic filtering rules of EgoActionBench. The benchmark contains 350 reference videos, with 4 target prompts for each video, resulting in 1,400 evaluation prompts. It includes 287 indoor scenes and 63 outdoor scenes, with 164 single-hand and 186 two-hand manipulation cases. In terms of data composition, approximately 60\% of the reference videos are sourced from real-world egocentric recordings (online videos and public manipulation datasets), while the remaining 40\% are generated by video diffusion models to cover rarer manipulation categories and ensure diversity. Here we provide additional details about camera-motion filtering, prompt construction, and manual quality control.

\paragraph{Camera-motion filtering.}
Egocentric videos often contain strong head or body motion, which can make action transfer evaluation ambiguous: a method may appear to fail because the reference observation is dominated by camera shake rather than hand-object interaction. To reduce this issue, we estimate camera poses with ViPE~\cite{huang2025vipe} for candidate real videos and compute frame-to-frame camera translation and rotation changes. Clips with large pose jumps, unstable pose estimates, or camera motion that dominates the hand-object region are removed before manual inspection. This filtering step keeps moderate egocentric viewpoint changes while excluding references where the manipulated object cannot be reliably tracked in the image plane.

\paragraph{Caption and prompt instructions.}
For each retained clip, we ask GPT-5 to produce an action-centric caption using the following instruction: \emph{Describe the egocentric manipulation action in one sentence. Focus on the actor's hand motion, the manipulated object, the contact transition, and the object's response. Do not describe camera motion, video quality, identity, or irrelevant background details.} Based on this source caption, we further ask GPT-5 to generate target prompts with the instruction: \emph{Rewrite the scene, object appearance, or background while preserving the same manipulation verb, contact order, and object response. Avoid copying reference-specific colors, textures, or scene details unless they are essential to the action.} This prompt design encourages evaluation cases that test action transfer rather than source appearance copying.

\begin{figure*}
  \centering
  \includegraphics[width=\linewidth]{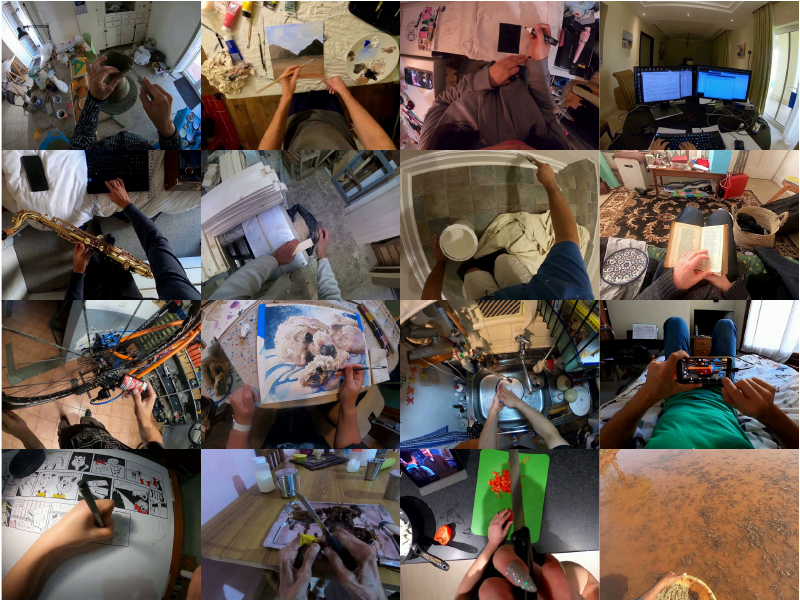}
  \caption{\textbf{More details about EgoActionBench.} EgoActionBench covers diverse egocentric manipulation scenarios and is annotated by scene type, and object category.}
  \label{fig:bench_supp}
\end{figure*}

\paragraph{Manual quality control.}
After automatic captioning and prompt rewriting, we manually review each benchmark sample. We reject prompts that change the action semantics, introduce physically incompatible objects, leak reference-specific appearance details, or describe interactions that are not visible in the source clip. We also check whether the target prompt is feasible for the reference action; for example, actions involving pouring, opening, or handover require objects with compatible affordances. The remaining samples are annotated with difficulty tags according to contact complexity, object motion magnitude, and whether the action involves one or two hands. These tags are used only for analysis and are not provided to the generation methods.

\section{More Results}

\begin{figure*}
 \centering
 \includegraphics[width=\linewidth]{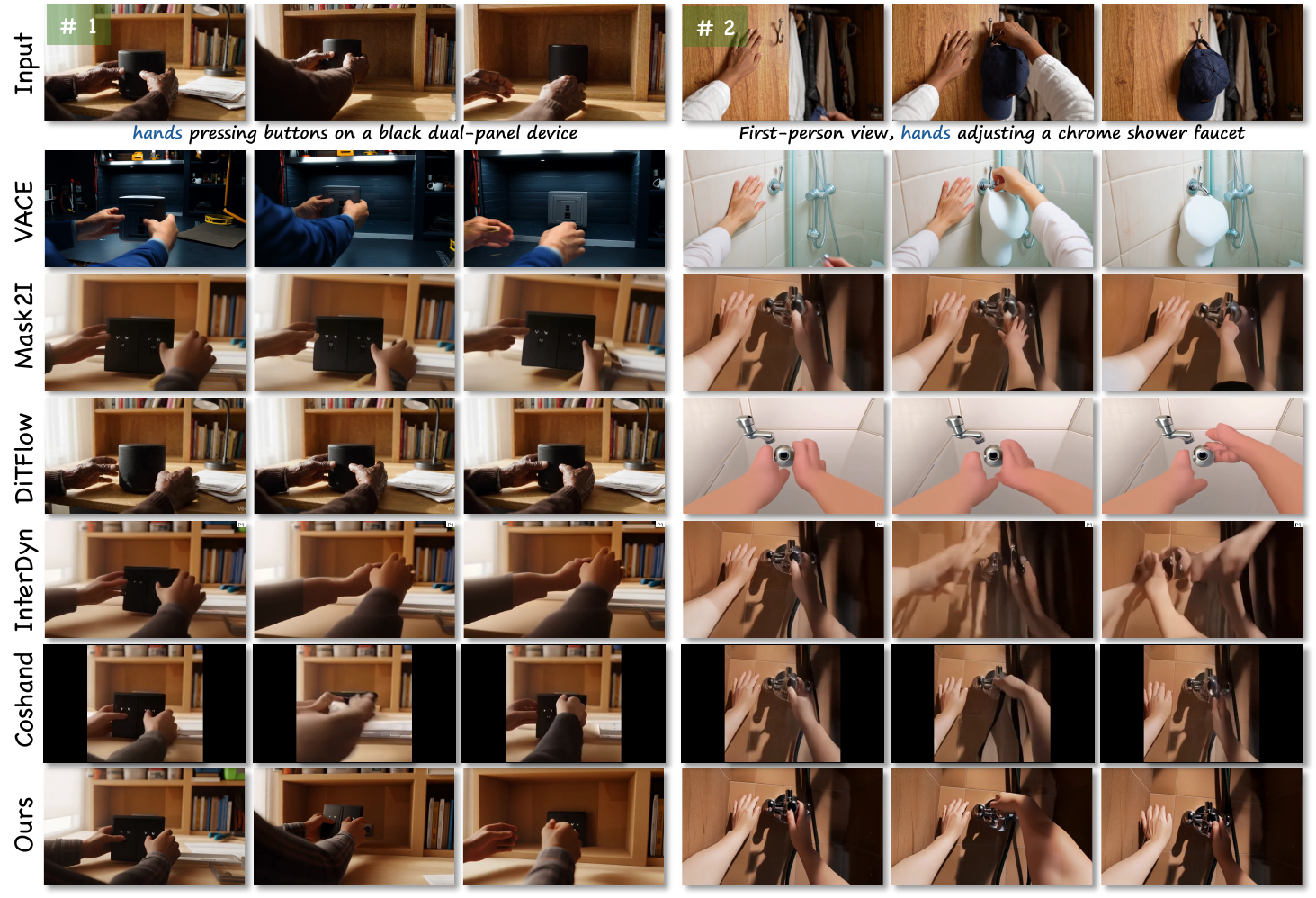}
 \caption{\textbf{More Comparison with Baseline.} Our method consistently produces more coherent action transfer and more plausible hand-object interactions than baselines, which often suffer from structural collapse, temporal inconsistency, or physically unrealistic interactions.}
 \label{fig:comp_supple}
\end{figure*}

\noindent \textbf{More Comparison with Baseline.} We provide more visual comparisons with baselines in Fig.~\ref{fig:comp_supple}. Our method consistently produces more coherent action transfer and more plausible hand-object interactions than baselines, which often suffer from structural collapse, temporal inconsistency, or physically unrealistic interactions.

\noindent \textbf{More Visual Results.} Figures~\ref{fig:gallery_supple1}, \ref{fig:gallery_supple2}, and \ref{fig:gallery_supple3} show more visual results of our method. The generated videos can be viewed in the demo video.

\begin{figure}
 \centering
 \includegraphics[width=\linewidth]{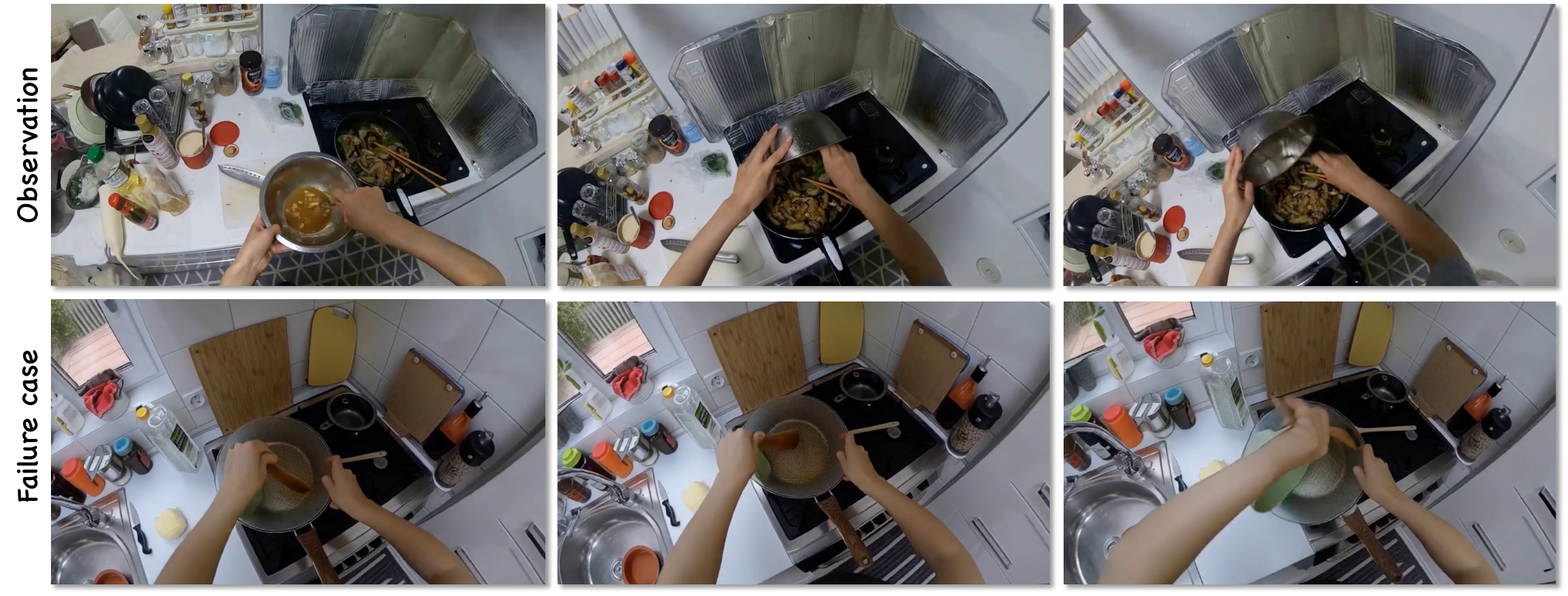}
 \caption{\textbf{Failure case.} Our method may struggle with actions involving very fast hand or object motion, which can lead to motion blur, temporal aliasing, or loss of hand-object correspondence. This limitation is primarily due to the base video diffusion model, which was not specifically trained to handle rapid egocentric dynamics and tends to produce temporally smoothed outputs when the inter-frame displacement is large.}
 \label{fig:failure_case}
\end{figure}
\begin{figure}[t]
 \centering
 \includegraphics[width=\linewidth]{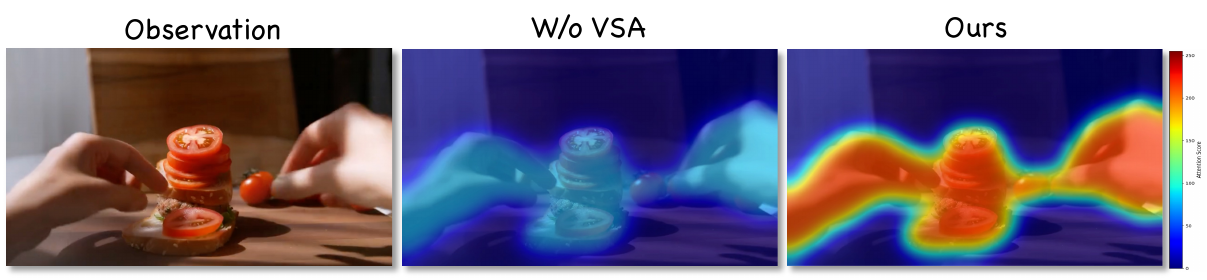}
 \caption{\textbf{Attention Visualization.} We visualize the cross-frame attention maps before and after enabling VSA during the early denoising stage. Without VSA, attention is dispersed across irrelevant regions, failing to capture coherent hand-object structure. With VSA enabled, the attention concentrates on interaction-relevant areas such as hand, object, and contact regions, indicating that the generation trajectory is better aligned with the reference structural layout. This confirms that VSA effectively anchors the hand-object skeleton through velocity-guided alignment, resulting in more structurally consistent attention patterns.}
\label{fig:attn}
\end{figure}

\section{Limitations and Future Work}
Although EgoACT achieves strong results on a wide range of egocentric manipulation scenarios, it still struggles with actions involving very fast hand or object motion. As shown in Fig.~\ref{fig:failure_case}, the generated videos may exhibit motion blur, temporal aliasing, or loss of hand-object correspondence in such cases. We attribute this limitation primarily to the base video diffusion model, which was not specifically trained to handle rapid egocentric dynamics and tends to produce temporally smoothed outputs when the inter-frame displacement is large. As video foundation models continue to advance in temporal resolution and motion modeling capacity, we expect this limitation to be naturally alleviated. In future work, we also plan to extend EgoACT to longer-horizon action sequences and multi-step task compositions, enabling more complex manipulation planning within the world simulation paradigm.

\section{Potential Social Impact}
Our work studies egocentric action transfer for controllable video world simulation. The potential positive impacts include more efficient simulation for embodied AI, robotics, and human-object interaction research, where generating diverse interaction videos can reduce the cost of collecting physical demonstrations and enable safer testing before real-world deployment. The method may also support content creation and instructional video synthesis when used with appropriate consent and attribution. At the same time, the ability to generate realistic egocentric manipulation videos may be misused to create misleading visual content or to synthesize actions that did not actually occur. Since our method builds on pre-trained video generation models, it may also inherit biases, failure modes, or unsafe generation behaviors from the underlying model and source data. To mitigate these risks, released examples and benchmark assets should be filtered for sensitive or harmful content, and generated videos should be clearly identified as synthetic when used outside research settings. We also recommend applying existing provenance, watermarking, or content moderation tools when deploying such systems in user-facing applications.

\begin{figure*}
 \centering
 \includegraphics[width=0.8\linewidth]{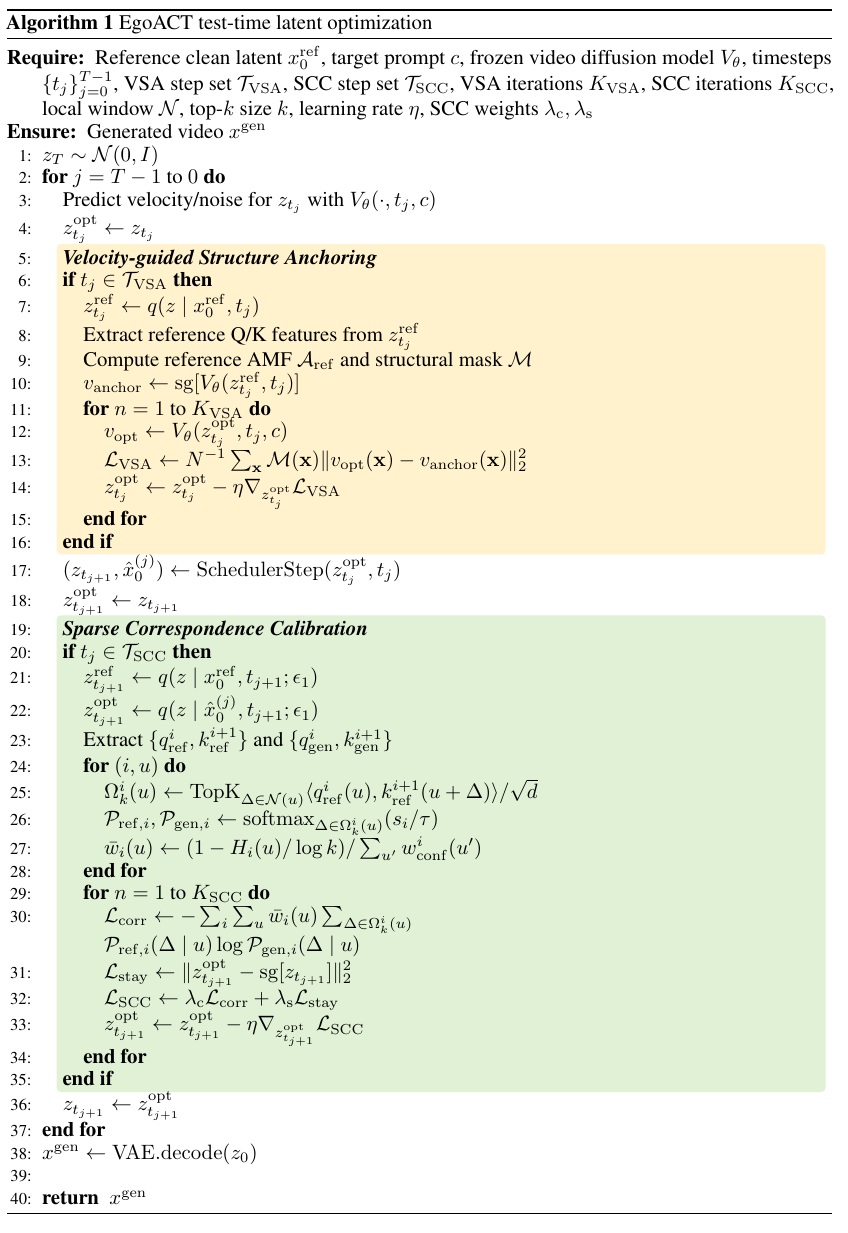}

 \label{fig:alg_supple}
\end{figure*}

\end{document}